\documentclass[sigconf,nonacm]{acmart}

\AtBeginDocument{%
  }

\copyrightyear{2026}
\acmYear{2026}
\setcopyright{cc}
\setcctype{by}
\acmConference[KDD '26]{Proceedings of the 32nd ACM SIGKDD Conference on Knowledge Discovery and Data Mining V.1}{August 09--13, 2026}{Jeju Island, Republic of Korea}
\acmBooktitle{Proceedings of the 32nd ACM SIGKDD Conference on Knowledge Discovery and Data Mining V.1 (KDD '26), August 09--13, 2026, Jeju Island, Republic of Korea}
\acmPrice{}
\acmDOI{10.1145/3770854.3780255}
\acmISBN{979-8-4007-2258-5/2026/08}

\usepackage{subfiles}
\usepackage{graphicx} 
\usepackage{subfig}
\usepackage{xcolor}
\usepackage{amsmath}
\usepackage{enumitem}

\usepackage[utf8]{inputenc}  
\DeclareUnicodeCharacter{FF0C}{,}  

\newtheorem{definition}{Definition}
\newtheorem{theorem}{Theorem}

\newtheorem{lemma}{Lemma}

\usepackage{algorithm}
\usepackage{algpseudocode}

\usepackage{multirow}
\usepackage{multicol}
\usepackage{makecell}

\usepackage{xspace}

\usepackage{mathtools, nccmath}

\newcommand{\algo}{\ensuremath{\text{PM-SFL}}\xspace}

\begin{document}

\title{Towards Privacy-Preserving and Heterogeneity-aware Split Federated Learning via Probabilistic Masking}





\author{Xingchen Wang}
\authornote{The first two authors contributed equally to this work.}
\orcid{0000-0003-1352-7445}
\affiliation{
  \institution{Purdue University}
  \city{West Lafayette}
  \state{IN}
  \country{USA}
}
\email{wang2930@purdue.edu}

\author{Feijie Wu}
\authornotemark[1]
\orcid{0000-0003-0541-1901}
\affiliation{
  \institution{Purdue University}
  \city{West Lafayette}
  \state{IN}
  \country{USA}
}
\email{wu1977@purdue.edu}

\author{Chenglin Miao}
\orcid{0000-0002-9646-7099}
\affiliation{
  \institution{Iowa State University}
  \city{Ames}
  \state{IA}
  \country{USA}
}
\email{cmiao@iastate.edu}

\author{Tianchun Li}
\orcid{0009-0007-7208-5524}
\affiliation{
  \institution{Purdue University}
  \city{West Lafayette}
  \state{IN}
  \country{USA}
}
\email{li2657@purdue.edu}

\author{Haoyu Hu}
\orcid{0009-0008-9585-7643}
\affiliation{
  \institution{Purdue University}
  \city{West Lafayette}
  \state{IN}
  \country{USA}
}
\email{hu943@purdue.edu}

\author{Qiming Cao}
\orcid{0000-0002-3329-3239}
\affiliation{
  \institution{Purdue University}
  \city{West Lafayette}
  \state{IN}
  \country{USA}
}
\email{cao393@purdue.edu}

\author{Jing Gao}
\orcid{0000-0002-1557-7553}
\affiliation{
  \institution{Purdue University}
  \city{West Lafayette}
  \state{IN}
  \country{USA}
}
\email{jinggao@purdue.edu}

\author{Lu Su}
\authornote{Lu Su is the corresponding author.}
\orcid{0000-0001-7223-543X}
\affiliation{
  \institution{Purdue University}
  \city{West Lafayette}
  \state{IN}
  \country{USA}
}
\email{lusu@purdue.edu}

\renewcommand{\shortauthors}{Xingchen Wang, et al.}

\begin{abstract}
  Split Federated Learning (SFL) has emerged as an efficient alternative to traditional Federated Learning (FL) by reducing client-side computation through model partitioning. However, exchanging of intermediate activations and model updates introduces significant privacy risks, especially from data reconstruction attacks that recover original inputs from intermediate representations. Existing defenses using noise injection often degrade model performance.  To overcome these challenges, we present \algo, a scalable and privacy-preserving SFL framework that incorporates Probabilistic Mask training to add structured randomness without relying on explicit noise. This mitigates data reconstruction risks while maintaining model utility. To address data heterogeneity, \algo employs personalized mask learning that tailors submodel structures to each client's local data. For system heterogeneity, we introduce a layer-wise knowledge compensation mechanism, enabling clients with varying resources to participate effectively under adaptive model splitting. Theoretical analysis confirms its privacy protection, and experiments on image and wireless sensing tasks demonstrate that \algo consistently improves accuracy, communication efficiency, and robustness to privacy attacks, with particularly strong performance under data and system heterogeneity.
\end{abstract}


  






\maketitle

\newcommand\kddavailabilityurl{https://doi.org/10.5281/zenodo.18090491}
\ifdefempty{\kddavailabilityurl}{}{
\begingroup\small\noindent\raggedright\textbf{Resource Availability:}\\
The source code of this paper has been made publicly available at \url{\kddavailabilityurl}.
\endgroup
}

\section{Introduction} \label{introduction}

Federated Learning (FL) has emerged as a promising paradigm for collaborative model training across decentralized devices while preserving data privacy \cite{mcmahan2017communication}. 
By exchanging model updates, FL addresses the challenge of limited data on individual devices. 
However, a practical barrier remains: many clients operate under tight computational constraints, making it difficult to train or store large models. 
Traditional model compression methods in FL, such as pruning \cite{li2021fedmask}, quantization \cite{gupta2022quantization} or knowledge distillation \cite{he2020group}, are generally unsuitable as they either require loading the entire model during training or lack a global model to support new clients.

To address this, Split Federated Learning (SFL) has been proposed as a resource-efficient variant of FL, where only part of the model (e.g., bottom layers) resides on the client side, while the remaining portion is hosted and trained on the server \cite{thapa2022splitfed, guo2023tree}. This reduces the memory and compute burden on clients and facilitates collaborative training. However, SFL introduces new privacy risks: clients must upload smashed data (intermediate activations) and updated local weights to the server, potentially exposing sensitive information. Our preliminary study in Section \ref{limitations} shows that standard SFL is vulnerable to data reconstruction attacks, which aim to recover original inputs from shared activations. Existing noise-based defenses offer limited protection unless strong noise is applied, which significantly degrades model performance. These findings reveal a fundamental trade-off between privacy and utility in SFL, raising a key question: \textit{Can we introduce randomness to obscure sensitive information while preserving model performance?}

To address this, we propose a novel framework, \algo, based on probabilistic mask training. On the client side, we train a learnable probabilistic mask for each parameter to identify sparse, structured subnetworks. Bernoulli sampling from these masks introduces structural randomness into the model, mitigating the risk of data reconstruction. Meanwhile, the resulting sparsity acts as implicit regularization, improving generalization by identifying important weights during training while filtering out redundant ones. This enables us to preserve model utility while enhancing privacy—without relying on strong noise injection. We rigorously formulate and instantiate this framework, and demonstrate its effectiveness both theoretically and empirically.

However, extending probabilistic masking to SFL is non-trivial due to data and system heterogeneity.
First, client data distributions are often highly non-IID \cite{li2019convergence, wu2024towards, wu2023anchor}, making it difficult to learn a global structure that generalizes across all clients. This necessitates personalized masks tailored to each client’s local data. Second, client devices differ significantly in compute, memory, and energy capacity \cite{kim2023depthfl, diao2020heterofl, wu2024fiarse, wu2023deterioration}, requiring adaptive model splitting where each client’s model depth matches its computational capacity.

To address these challenges, \algo incorporates dedicated mechanisms for both data and system heterogeneity. 
To address data heterogeneity, we propose mask-based partial model personalization, where a subset of masks is always kept for local training to preserve personalization. Each client measures how much the sampling behavior of its probabilistic masks changes during local updates. Masks with large changes, indicating strong local–global disagreement, are excluded from global aggregation, while the remaining masks are aggregated. This separation enables better local adaptation and accelerates global convergence.
To tackle system heterogeneity, we adopt adaptive model splitting, allowing clients to train only as deep as their resources permit. Since fewer clients train the deeper layers, those layers receive far fewer updates and become undertrained. To address this, we introduce a probabilistic-masking-based knowledge compensation mechanism that recovers missing updates from shallow clients in a privacy-preserving manner. 
Taken together, \algo jointly addresses privacy and heterogeneity across clients, adapting to diverse client requirements while enabling effective global model construction.

\noindent\textbf{Contributions.} The contributions of this paper can be highlighted from the following perspectives:
\begin{itemize}[leftmargin=1em,topsep=0pt]
    \item \textbf{Algorithmically}, we propose \algo, a novel SFL framework that leverages probabilistic mask training to enhance privacy with minimal utility loss. To support system and data heterogeneity, we further introduce personalized mask training and layer-wise knowledge compensation, enabling efficient and customized learning on resource-constrained clients.
    
    \item \textbf{Theoretically}, we analyze the privacy protection provided by \algo, showing enhanced robustness against data reconstruction attacks and improved differential privacy guarantees compared to traditional noise-injection methods.
    
    \item \textbf{Empirically}, we validate \algo on image and wireless sensing datasets using ResNet and Transformer backbones. \algo outperforms baselines in accuracy, communication efficiency, and resilience to privacy attacks. Its advantages are pronounced under data and system heterogeneity.
\end{itemize}

\section{Background} \label{bacground}


\begin{table}[t] 
\footnotesize
\centering
\renewcommand{\arraystretch}{1.2}
\caption{Summary of notation used in this paper.}
\resizebox{\linewidth}{!}{
\begin{tabular}{cl}
\hline\rule{0pt}{2.2ex} 
$N, i$ & total number, index of \emph{clients} \\
$\mathcal{K}, K$ & set, number of \emph{sampled clients}\\
$T, t$ & number, index of \emph{communication rounds} \\
$R, r$ & number, index of \emph{local update iterations} \\ 
$\mathbf{w}_b, \mathbf{w}_p$ & split model at the client side and the server side \\
$\theta_t$ & probabilistic mask in round $t$ \\
$\mathbf{s}_{t, r}^{(k)}$ & score mask of client $k$ in round $t$ and iteration $r$\\
$\mathbf{M}_{k}$ & binary mask of client $k$ on top of the latest score mask\\
$\mathcal{Q}_k, \nabla\mathcal{Q}_k$ & smashed data and their gradients of client $k$ \\
$\mathbf{I}_k$ & local binary indicator of client $k$ \\
$\mathcal{K}_t^l$ & set of \emph{sampled clients} with layer $l$ in round $t$ \\
$(\theta_t^l)^{sl}, (\theta_t^l)^{client}$ & \makecell[l]{probabilistic mask at the server, client side in round $t$\\with layer $l$}\\
\hline
\end{tabular}
}
\vspace{10pt}
\label{table:notation}
\end{table}

\subsection{Split Federated Learning}  \label{sec:SFL}

Federated Learning (FL) enables collaborative model training across distributed clients without sharing raw data. Consider $N$ clients, where each client $i$ holds a local dataset $D_i = \{(x_i^{(j)}, y_i^{(j)})\}_{j=1}^{N_i}$. Let $\mathbf{w} \in \mathbb{R}^d$ be the model, and define the empirical loss for client $i$ as $F_i(\mathbf{w}) = \frac{1}{N_i} \sum_{(x, y) \in D_i} \ell(f_{\mathbf{w}}(x); y)$. The global optimization objective is:
\begin{equation}
    \min_{\mathbf{w} \in \mathbb{R}^{d}} F(\mathbf{w}) \triangleq \frac{1}{N} \sum_{i=1}^{N} F_i(\mathbf{w})
\end{equation}

In each round, clients train locally and upload updates to a server for aggregation, which refines the global model and redistributes it. However, full-model training imposes high memory and computation costs on resource-limited clients. To address this, model compression (e.g., quantization~\cite{gupta2022quantization}, sparsification~\cite{li2021fedmask}) has been explored, though often at the expense of accuracy.

\begin{figure}[t]
    \centering
    \includegraphics[width=0.85\linewidth]{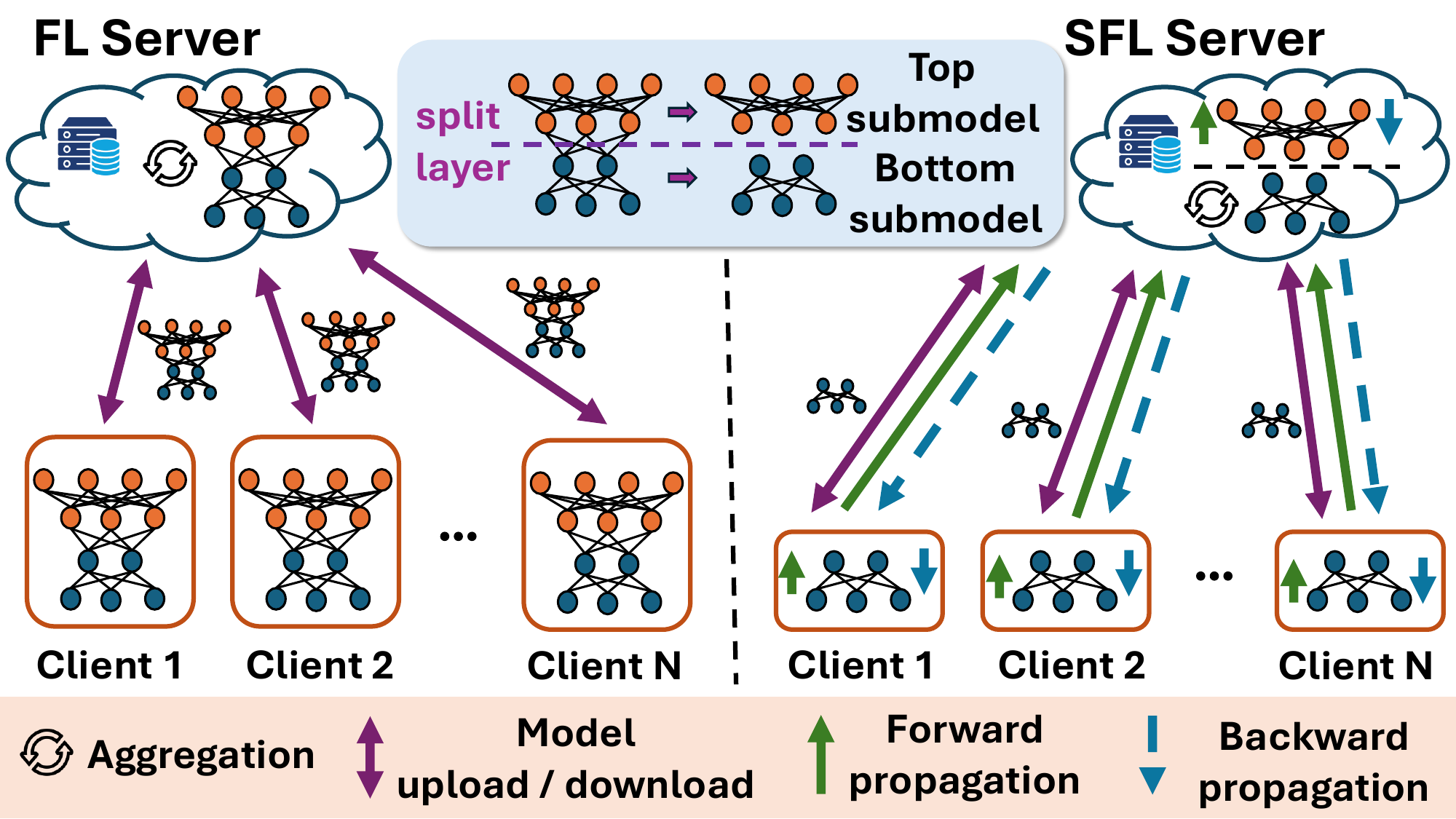} 
\vspace{1mm}
    \caption{Illustration of typical FL and SFL.}
    \label{fig:FL_SFL}
\vspace{1mm}
\end{figure}

Split Federated Learning (SFL) alleviates these challenges by splitting the model into a lightweight bottom submodel $\mathbf{w}_b \in \mathbb{R}^{d_b}$ on the client and a top submodel $\mathbf{w}_p \in \mathbb{R}^{d_p}$ on the server, as illustrated in Figure \ref{fig:FL_SFL}. Clients compute intermediate activations (smashed data) using $\mathbf{w}_b$, which are sent to the server for the remaining forward/backward passes. The global objective becomes:
\begin{equation} \label{eq:SFL}
    \min_{\mathbf{w}_b, \mathbf{w}_p} F(\mathbf{w}_b, \mathbf{w}_p) \triangleq \frac{1}{N} \sum_{i=1}^{N} \frac{1}{|D_i|} \sum_{(x,y) \in D_i} \ell(f_{\mathbf{w}_p}(f_{\mathbf{w}_b}(x)); y)
\end{equation}

This design significantly reduces client-side computation and memory usage, making federated learning feasible for resource-constrained environments. The server can also aggregate bottom submodels to facilitate global consistency.

\subsection{Data Reconstruction Attack} \label{sec:data_reconstruction_attck}

Data reconstruction attacks aim to recover clients’ raw training data from information shared during the training process, posing a significant threat to privacy in federated systems. To evaluate the privacy risks in split federated learning (SFL), we consider a representative reconstruction attack inspired by Deep Leakage from Gradients (DLG) \cite{zhu2019deep}, assuming an honest-but-curious server as the adversary. 
In standard FL, gradients are derived from model updates sent from clients to the server. DLG initiates reconstruction by creating dummy inputs and labels, then optimizing them to produce dummy gradients that closely match the actual gradients, ultimately recovering the original input data. 
In SFL, clients transmit not only parameter updates but also intermediate representations—referred to as smashed data—to the server. This additional information simplifies the reconstruction process, as both gradients and model outputs can be jointly exploited to more effectively approximate the original client data.



\section{Limitations of SFL and Current Defense} \label{limitations}


\begin{figure}[t]
    \centering
    
    \includegraphics[width=0.95\linewidth]{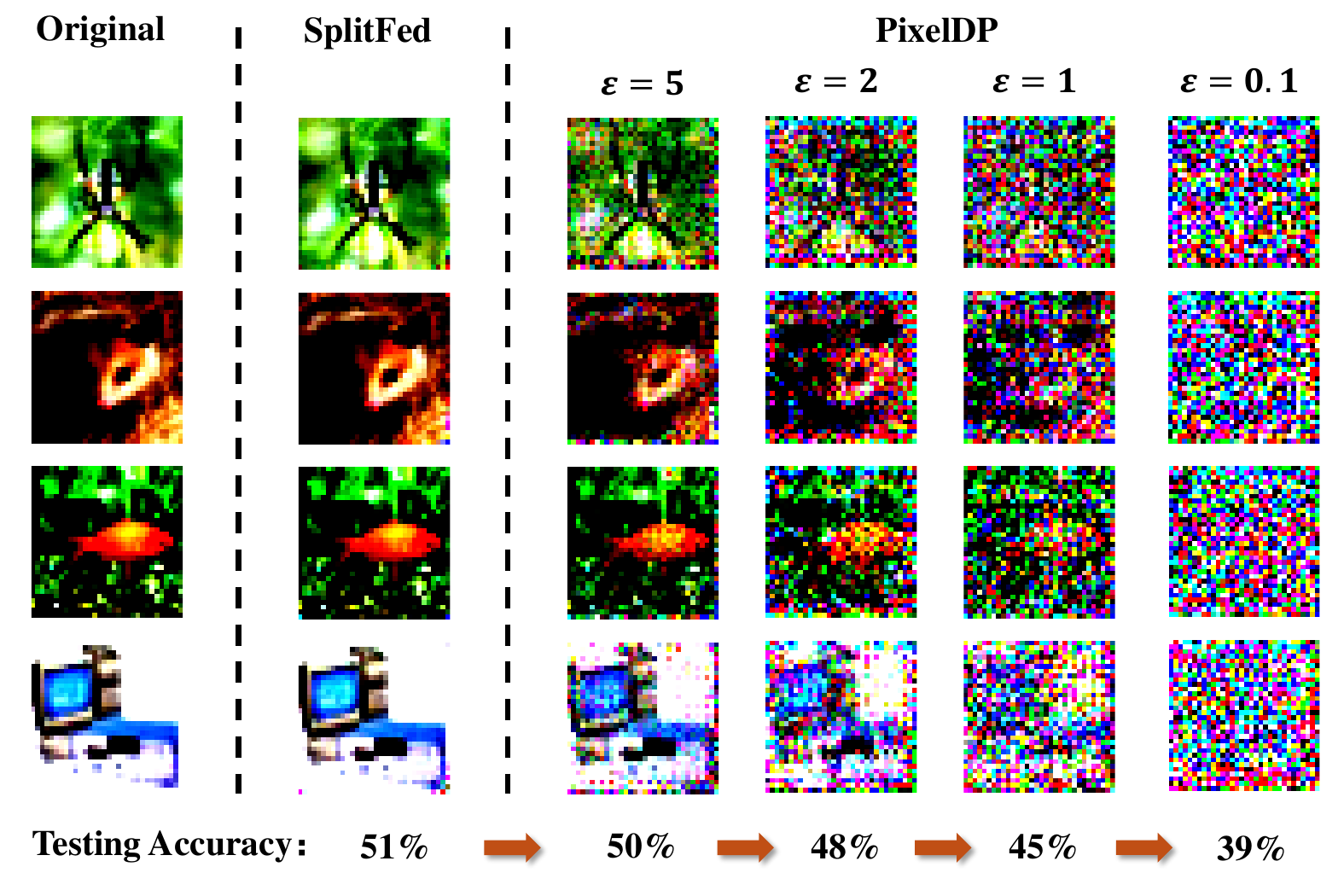} 
    
    \caption{Visualization of reconstructed data  for vanilla SFL (SplitFed) and noise-injection-based variants (PixelDP) under different privacy budgets on CIFAR-100 using ResNet-18.}
    \label{fig:reconstruction_attack}
\vspace{1mm}
\end{figure}

While SFL significantly alleviates the computational burden on the client side, the process of uploading smashed data from the split layer and sharing the updated local bottom submodel with the server introduces risks to data privacy. 
To assess the privacy vulnerabilities of split federated learning, we perform a data reconstruction attack described in Section \ref{sec:data_reconstruction_attck}.
Figure \ref{fig:reconstruction_attack} illustrates that the data reconstruction attack can achieve complete reconstruction in SFL. This demonstrates a substantial privacy risk in vanilla SFL (SplitFed), highlighting the urgent need for effective protection mechanisms to secure data privacy.

Noise injection is commonly adopted for protecting data privacy in SFL \cite{thapa2022splitfed}. This approach involves adding calibrated noise to smashed data, as in PixelDP \cite{lecuyer2019certified}, and to local model updates, as in DP-SGD \cite{abadi2016deep}, before transmission from the client. The amount of injected noise is determined by the chosen privacy budget. 
To evaluate the effectiveness of this defense mechanism, we conduct the same data reconstruction attack on SFL variants employing noise injection under different privacy budgets. Following the privacy study design in \cite{thapa2022splitfed}, we fix a high privacy budget for local model update protection at $\epsilon = 5$ and vary the privacy budget for smashed data protection from $\epsilon = 5$ to $\epsilon = 0.1$.
Our results in Figure~\ref{fig:reconstruction_attack} reveal that traditional noise injection provides limited protection when the privacy budget is high, leaving sensitive information vulnerable. Only when the privacy budget is reduced to $\epsilon = 0.1$, the injected noise sufficiently obfuscates the data to prevent effective reconstruction. However, this strong protection comes at a significant cost: model accuracy drops from $51\%$ (without protection) to $39\%$.

These findings reveal a fundamental trade-off in noise injection protection: injecting noise enhances privacy but degrades model utility. This raises the question: \textit{can we introduce randomness in a way that both protects privacy and preserves model performance?}
Our proposed solution, probabilistic mask training, addresses this challenge. It learns a probabilistic mask for each parameter, guiding the selection of an optimal sparse subnetwork via Bernoulli sampling. This sampling introduces structural randomness into the model, which can obscure sensitive client information during communication with the server. Meanwhile, it filters out less important parameters, reducing overfitting and enhancing generalization. Thus, this approach enhances privacy while maintaining accuracy.

\section{Methodology} \label{method}



Our goal is to refine SFL by enhancing privacy protection while minimizing the performance degradation often associated with traditional privacy-preserving techniques. Simultaneously, we aim to deliver an optimal customized model for each client, tailored to their data and computational resources, while also leveraging shared knowledge to train an effective global model.

In this section, we begin by detailing how probabilistic mask training is applied in the SFL setting (Section \ref{sec:prob_mask}). We then present two extensions to the core training method to address data heterogeneity (Section \ref{sec:per_mask}) and system heterogeneity (Section \ref{sec:knoeledge_compensation}). Finally, we summarize the complete \algo framework in Algorithm \ref{algo:framework}.

\subsection{Training with Probabilistic Mask} \label{sec:prob_mask}


Instead of using traditional weight training methods for bottom submodel introduced in Section \ref{sec:SFL}, we freeze the weights $\mathbf{w}_b \in \mathbb{R}^{d_b}$ initialized with Kaiming Normal distribution \cite{he2015delving}, which helps control the variance of the neurons' output to prevent the vanishing or exploding of activation values (or if the pre-trained model is available, we initialize the weights with the pre-trained ones). Subsequently, we focus on learning how to effectively sparsify the network to optimize model performance by finding the optimized binary mask $\mathbf{M}\in\{0,1\}^{d_b}$. 

\begin{figure} [t]
    \centering
    \includegraphics[width=0.85\linewidth]{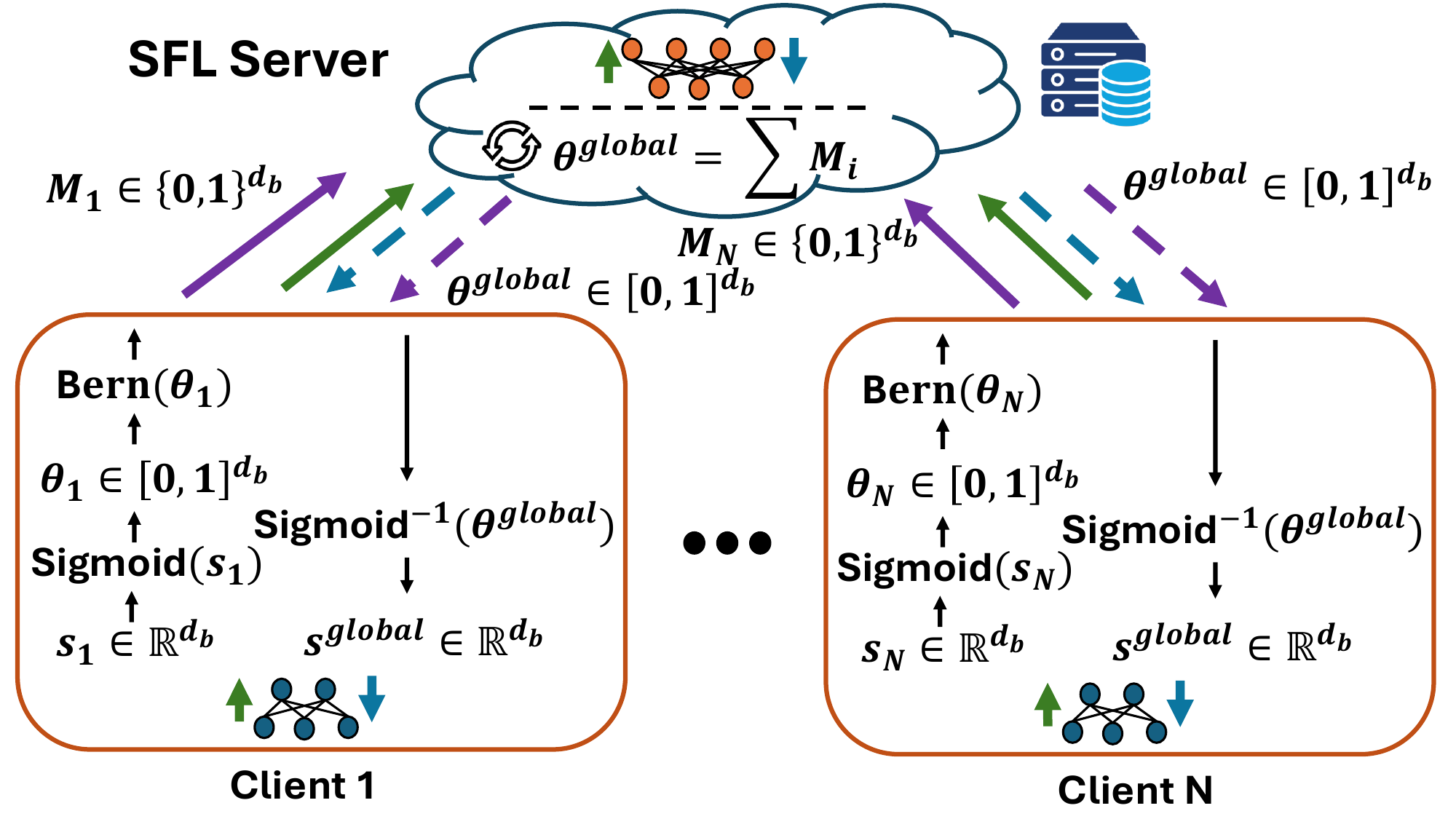} 

    \caption{Probabilistic Mask Training in SFL.}
    \label{fig:prob_mask}
\end{figure}

By introducing probabilistic masks:  $\theta\in[0,1]^{d_b}$, which describe the probability of parameter being kept or pruned, randomness is injected into model sparsification: $\mathbf{M} \sim \text{Bern}(\theta)$, making the local model unpredictable and thereby enhancing privacy protection. 
Since $\theta \in [0, 1]^{d_b}$, the problem becomes a constrained optimization. This can be solved by clipping before taking a Bernoulli sampling, but it may introduce bias and further slow convergence and reduce accuracy. Instead, we use a score mask $\mathbf{s} \in \mathbb{R}^{d_b}$ that has unbounded support to generate $\theta$ via a sigmoid function $\mathbf{\theta} = \sigma(\mathbf{s})$, illustrated in Figure \ref{fig:prob_mask}. Then, the objective function is formulated as:
\begin{equation}
\begin{aligned}
    \min_{\mathbf{s} \in \mathbb{R}^{d_b}, \mathbf{w}_p\in \mathbb{R}^{d_p}} F\left(\mathbf{w}_{b} \odot \mathbf{M}, \mathbf{w}_p\right) & \triangleq \frac{1}{N} \sum_{i=1}^{N} F_i(\mathbf{w}_{b} \odot \mathbf{M}, \mathbf{w}_{p}), 
\end{aligned}
\end{equation}
This objective ensures the extracted submodel achieves performance comparable to the one trained with the weights \cite{frankle2018lottery, zhou2019deconstructing}.

Given the non-trivial nature of the optimization, we propose the following steps for probabilistic mask-based SFL.
After initializing the client’s local model, we repeat the following steps $1$ to $3$ for $T$ communication rounds. At each communication round $t \in \{0, 1, \dots, T-1\}$:

\noindent\textbf{Step $1$ (Participants Sampling):}\quad The server samples a set $K_t$ of $|K_t| = K$ participants (out of the total $N$ clients) without replacement to participate in the training process. The server broadcasts the probabilistic mask $\mathbf{\theta}_t$ to the selected participants.

\noindent \textbf{Step $2$ (Local Updates):} After the selected client $k \in K_t$ receives the probabilistic mask $\theta_t$ from the server, it initializes the score mask to $\mathbf{s}_{t,0}^{(k)} = \sigma^{-1}(\mathbf{\theta}_{t})$, where $\sigma^{-1}(\cdot)$ is an inverse function of the Sigmoid function. Since a communication round consists of multiple local updates, the client and the server collaboratively implement the following steps $2.1$ to $2.3$ for $R$ iterations. At each iteration $r \in \{0, \dots, R-1\}$, 

\noindent \textbf{Step $2.1$ (Local Forward):}\quad The selected client $k \in K_t$ 
samples a local mask $\mathbf{M}_k \sim \text{Bern}(\sigma(\mathbf{s}_{t, r}^{(k)}))$ and samples a mini-batch $\mathcal{B}_k$ and computes the smashed data $\mathcal{Q}_k = \{(f_{\mathbf{w}_b \odot \mathbf{M}_k}(x), y)\}_{(x, y) \in \mathcal{B}_k}$. Then, the client forwards the smashed data $\mathcal{Q}_k$ to the server. 

Clients can choose whether to share labels with the server \cite{thapa2022splitfed}. Prior work protects labels in SFL by keeping them local \cite{han2021accelerating, he2020group} or adding perturbations \cite{yang2022differentially}. These orthogonal approaches can be readily integrated into our method.

\noindent \textbf{Step $2.2$ (Gradient Computation on Server):}\quad The server receives the smashed data from all selected participants $K_t$, denoted by $\mathcal{Q} = \cup_{k \in K_t} \mathcal{Q}_k$. With the collected smashed data, the server will update the server model $\mathbf{w}_p$ using SGD. 
In the meantime, the server computes the gradient with respect to the smashed data for each client, i.e., $\nabla \mathcal{Q}_k = \left\{\frac{\partial \ell(f_{\mathbf{w}_p} (q); y)}{\partial q}\right\}_{q \in \mathcal{Q}_k}$, which is transmitted back to corresponding clients.

\noindent \textbf{Step $2.3$ (Local Backward):} After receiving the gradient w.r.t. the smashed data, client $k \in K_t$ updates its local score $\mathbf{s}_{t, r}^{(k)}$ via
\begin{align}\label{eq:local_update}
    \mathbf{s}_{t, r+1}^{(k)} = \mathbf{s}_{t, r}^{(k)} - \frac{1}{|\mathcal{Q}_k|} \sum_{q \in \mathcal{Q}_k, \nabla q \in \nabla \mathcal{Q}_k} (\nabla q)^{\top} \left(\frac{\partial q}{\partial \mathbf{s}_{t, r}^{(k)}}\right),
\end{align}
where 
$\frac{\partial q}{\partial \mathbf{s}_{t, r}^{(k)}} = \frac{\partial f_{\mathbf{w}_b \odot \mathbf{M}_k}(x)}{\partial \mathbf{w}_b \odot \mathbf{M}_k} \odot \mathbf{w}_b \odot \sigma(\mathbf{s}_{t, r}^{(k)}) \odot \sigma'(\mathbf{s}_{t, r}^{(k)})$
and $\sigma'(\cdot)$ is the derivative of $\sigma(\cdot)$, 
which follows \citet{isik2022sparse} and makes use of the straight-through estimator \cite{bengio2013estimating} by assuming that the Bernoulli sampling operation is differentiable during backward propagation. 
A detailed analysis is provided in the Appendix \ref{sec:backward_derivation}.


\noindent \textbf{Step $3$ (Mask Aggregation):} After $R$ iterations of local updates, each selected client $k \in K_t$, instead of sending the probabilistic mask $\theta_{t, R}^{(k)} = \sigma(\mathbf{s}_{t, R}^{(k)})$ to the server, samples a binary mask $\mathbf{M}_k$ for aggregation, i.e., $\mathbf{M}_k \sim \text{Bern}(\theta_{t, R}^{(k)})$. Therefore, the server updates the global $\mathbf{\theta}_{t+1}$ as 
\begin{align}\label{eq:mask_agg}
    \mathbf{\theta}_{t+1} = \frac{1}{K} \sum_{k \in K_t} \mathbf{M}_k.
\end{align}

In principle, the server aggregates the probabilistic masks $\boldsymbol{\theta}_k$ via $\theta = \frac{1}{K} \sum_{k \in K_t} \boldsymbol{\theta}_k$ and shares the global mask with clients each round. We have clients upload sampled binary masks $\mathbf{M}_k \sim \text{Bern}(\boldsymbol{\theta}_k)$ for two reasons: (1) it reduces communication to less than one bit per parameter \cite{he2023gluefl, wu2022sign}, and (2) it mitigates data reconstruction risks in extreme cases, such as when a client holds only a single sample as analyzed in Appendix \ref{sec:leakage_analysis}.
Moreover, the estimator $\hat{{\theta}} = \frac{1}{K} \sum_{k \in K_t} \mathbf{M}_k$ is unbiased and has a provably bounded estimation error \cite{isik2022sparse}:
\begin{equation}
    \mathbb{E}_{\mathbf{M}_k \sim \text{Bern}(\boldsymbol{\theta}_k), \forall k \in K_t}  
\left[ \left\| \hat{\mathbf{\theta}} - \mathbf{\theta} \right\|_2^2 \right] 
\leq \frac{d}{4K}.  
\end{equation}

\subsection{Data-aware Mask Personalization} \label{sec:per_mask}



To address data heterogeneity and enhance personalization, we propose partial model personalization within our probabilistic mask framework. In this method, each client selects a subset of masks to be consistently trained locally, while the remaining masks are aggregated in every communication round. The personalized masks are tailored to capture the unique characteristics of each client's local data distribution, whereas the non-personalized masks enable the sharing of common knowledge. This strategy balances local adaptation with the effectiveness of the global model.

Specifically, we design a local binary indicator $\mathbf{I}_k \in \{0,1\}^{d_b}$ to specify whether each probabilistic mask should be personalized. Each client uses this indicator to partition its masks into two groups: personalized and non-personalized. The primary challenge lies in effectively determining the values of this indicator.


\noindent\textbf{Personalized Mask Selection.}\quad Building on insights from \cite{tamirisa2024fedselect}, we hypothesize that after several communication rounds, during which clients converge on an initial global consensus, then the local updates in the following rounds are primarily driven by the discrepancies between local and global data distributions. Therefore, masks undergoing the most significant changes during local training should be personalized, while those with minimal changes should be globally aggregated. 
To achieve this, we analyze the element-wise difference between the client’s masks  before and after local training:
$|\Delta\mathbf{\theta}_k| = |\sigma(\mathbf{s}_k) - \sigma(\mathbf{s'}_k)|$, where $\mathbf{s'}_k$ and $\mathbf{s}_k$ represent the score mask before and after one round of local training.

Unlike \cite{tamirisa2024fedselect}, our training target is probabilistic masks rather than parameters. In this context, the magnitude of a probabilistic mask carries a distinct meaning: values closer to $1$ or $0$ indicate greater certainty about whether the corresponding parameters should be retained or removed from the subnetwork. Consequently, masks with values above $0.5$ or below $0.5$ express opposing views regarding parameter selection in the subsequent sampling process. We call those transitioning across the $0.5$ threshold (i.e., from greater than $0.5$ to less than $0.5$, or vice versa) as Disagree Group $|\Delta\theta|_{dis}$ and those not across the threshold as Agree Group $|\Delta\theta|_{agr}$.
We prioritize the search space for personalized masks to Disagree Group, as such transitions indicate a divergence between local and global preferences on parameter selection, making them more suitable for personalization. Additionally, by excluding these divergent updates from global aggregation, the proposed method also accelerates global convergence. 
Personalized mask selection is performed incrementally, with the set of personalized masks growing gradually over communication rounds.

\noindent\textbf{Heterogeneity-aware Aggregation.}\quad In each communication round, after clients perform their local updates, they send the updated mask and the personalization indicator to the server. Given the potential heterogeneity of these indicators across different clients, a straightforward averaging of the updated masks is not feasible. Instead, for each probabilistic mask, aggregation is performed only among those clients that overlap on the specific mask: 
$\sum^{K}_{i=1}  \mathbf{M}_{i} \odot \neg\mathbf{I}_{i}$, where $\neg$ denotes the binary inversion operator, ensuring that only non-personalized masks are aggregated.
This sum is then divided by the number of clients contributing to the aggregation for this mask. Consequently, the aggregated mask is computed as follows:
\begin{equation} \label{eq:aggregation}
    \theta = \left(\sum^{K}_{i=1}  \mathbf{M}_{i} \odot \neg\mathbf{I}_{i}\right) \oslash \left(\sum^{K}_{i=1} \neg\mathbf{I}_{i}\right)
\end{equation}
where $\oslash$ denotes element-wise division.


 

\subsection{Layer-wise Knowledge Compensation} \label{sec:knoeledge_compensation}

Computational resources vary across clients, making a fixed bottom submodel impractical. Enforcing a uniform bottom submodel either restricts its size to accommodate the least capable client—exacerbating vulnerability to data leakage—or excludes resource-limited clients from participation. To address this, we propose adaptive model splitting, where each client’s local model depth is adaptively determined based on its computational capacity. This approach maximizes the client-side model size while enabling resource-constrained clients to participate in collaborative learning. 

However, our preliminary study revealed that simply adjusting the depth of the local model and averaging the overlapping layers on the server significantly degrades model performance. This degradation arises when clients with limited computational resources can only locally update submodels with shallow layers, preventing their participation in the aggregation of deeper layers. As a result, the aggregated deeper layers suffer from insufficient training data. Actually, for these resource-limited clients, deeper layer learning on their local data occurs in the split learning step on the server side. To fill the gap in training data during the aggregation phase, it is essential to compensate for this missing information. 

Fortunately, by integrating probabilistic masking training, we enhance data privacy, even when top submodel training and bottom submodel aggregation are handled by a single server. This enables top model updates related to clients with only shallow layers to participate in aggregation without compromising privacy. Specifically, we modify the training process of the top model to integrate probabilistic masking, allowing the masks learned on the top submodel to be directly used in aggregation.

The server performs aggregation layer-by-layer; for each layer, it calculates a weighted average of updates from both clients and the top submodel on the server.  The weights are determined by the ratio of the number of clients contributing to layer $l$ relative to the number of resource-constrained clients absent from the aggregation for layer $l$, ensuring that all layers are adequately trained.
At communication round $t$, we define the set of training participants that include the layer $l$ in their local submodel as $\mathcal{K}^l_t$. The aggregation for knowledge compensation is then conducted as follows:
\begin{equation}
    \theta^{l}_{t} = (1-\frac{|\mathcal{K}^l_t|}{|\mathcal{K}_t|})(\theta^{l}_{t})^{sl} + \frac{|\mathcal{K}^l_t|}{|\mathcal{K}_t|}(\theta^{l}_{t})^{client}
\end{equation}
where $(\theta^{l}_{t})^{sl}$ represents the update from the server-side split learning on layer $l$ at round $t$ while $(\theta^{l}_{t})^{client}$ denotes the aggregated update from the clients at round $t$, as shown in Equation \ref{eq:aggregation}.



\noindent\textbf{Summary.}\quad Algorithm \ref{algo:framework} summarizes the full \algo framework.

\begin{algorithm}[t]
\caption{\algo. There are $N$ clients with index of $i$, $C$ is the ratio of clients selected every round.}
\label{algo:framework}
\begin{algorithmic}[1]
\State \texttt{/* Adaptive Model Splitting */}
\State Server initializes global model with weights $w$; split the model according to each client's local computational resources, and distribute the corresponding bottom submodel $w_{b,i}$ to client $i$
\For{each round $t = 1, 2, \ldots$}
    \State $K \gets \max(C \cdot N, 1)$; $\mathcal{K}_t \gets$ (random set of $K$ clients)
    \State Server broadcasts global prob mask to each selected client
    \For{each client $k \in \mathcal{K}_t$ in parallel}

        \State \texttt{/* Probabilistic Mask-based Split Learning */}
        \For{each local epoch}
            \State Local forward with sampled mask; send smashed data to server
            \State Server computes gradient and updates top submodel weights; returns smashed data gradient back
            \State Local backward and update probabilistic mask
        \EndFor 
        
        \State \texttt{/* Personalized Mask Selection */}
        
        \State Calculate the probabilistic mask change $|\Delta\theta|$
        \State Split the mask change into Disagree Group and Agree Group: $(|\Delta\theta|_{dis}, |\Delta\theta|_{agr})$
        \State Prioritize $|\Delta\theta|_{dis}$ to select top $r$ values in $|\Delta\theta|$
        \State Send updated non-personalized binary masks to server
    \EndFor

    \State \texttt{/* Heterogeneity-aware Aggregation \& Layer-wise Knowledge Compensation */}
    
    \For{each model layer $l$}
        \State Aggregate clients' heterogeneous updates
        \State Compensate aggregation from server-side split learning
    \EndFor
\EndFor
\end{algorithmic}
\end{algorithm}

\section{Privacy Protection Analysis} \label{privacy_protection}


The design of probabilistic mask training introduces structural randomness into the local submodel, enhancing privacy. We examine its impact on data privacy below.

\noindent \textbf{Protection Against Data Reconstruction Attack.}\quad
Since the client sends smashed data to the server, the risk of input reconstruction arises. To assess this threat, we analyze the reconstruction error as follows:
\begin{theorem} \label{theorem:data_reconstruction}
Suppose there is a 
deep neural network with the parameters of $\mathbf{w}_b \in \mathbb{R}^{n \times m}$, i.e., $f_{\mathbf{w}_b}: \mathbb{R}^n \rightarrow \mathbb{R}^m$, and $\mathbf{w}_b$ is a non-singular matrix. 
The raw input $x \in \mathbb{R}^n$ with a randomly generated mask $M$ yields a smashed data $f_{\mathbf{w}_b \odot M}(x)$, where the mask $M$ is generated by the Bernoulli sampling with parameter $p \in [0, 1]^{n \times m}$.
As the server can witness the smashed data $f_{\mathbf{w}_b \odot M}(x)$ and the probability $p$, the server can attempt to reconstruct the raw data ($\hat{x}$ denotes the reconstructed data), where the expected reconstruction error is:
\begin{align}
\mathbb{E}\|x-\hat{x}\| \ge \sum_{|\hat{M} - M| \ne 0} P(\hat{M})\,\frac{\lambda_{\min}(\mathbf{w}_b \odot (\hat{M}-M))}{\lambda_{\max}(\mathbf{w}_b \odot \hat{M})}\|x\|.
\end{align}
$\lambda(\cdot)$ indicates the singular value of a matrix. $\hat{M}$ is a randomly generated mask with a probability of $P(\hat{M})$. 
\end{theorem}

This analysis highlights the lower bound for the distance between real and speculative samples. The speculative mask $\hat{M}$ leads $\mathbf{w}_b \odot (\hat{M} - M)$ to a singular matrix, making the minimum singular value nonzero, preventing the error from vanishing. More importantly, the bound scales with both the norm of the input $\|x\|$ and the probability of generating a mask that introduces sufficient randomness—quantified by $\sum_{|\hat{M} - M| \ne 0} P(\hat{M})$. This reveals a key strength of the probabilistic masking: it induces structural uncertainty that fundamentally limits the attacker’s ability to accurately reconstruct the original input. 
The proof is provided in Appendix \ref{proof_theorem1}

\noindent \textbf{Enhancement of Differential Privacy (DP).}\quad
While the above theorem provides a lower bound guarantee on data reconstruction error, we further analyze the proposed method through the lens of Differential Privacy (DP) \cite{dwork2006differential, dwork2014algorithmic}, which offers a more rigorous and general mathematical framework for quantifying privacy guarantees against a broader range of attacks. Formally, its definition is given as follows:
\begin{definition}[Adjacent Datasets]
Two datasets $D, D' \in \mathcal{D}$ are called adjacent if they differ in at most one data sample. 
\end{definition}
\begin{definition}[$(\epsilon, \delta)$-DP]\label{def:dp}
A randomized mechanism $\mathcal{M}$ is $(\epsilon,\delta)$-differentially private if for any neighboring datasets $D$ and $D'$ that differ by one record, and any subset $\mathcal{O}$ of the output domain of $A$, we have the following.
\begin{align}\label{eq:dp-def}
    \Pr[\mathcal{M}(D)\in\mathcal{O}] \leq \exp(\epsilon) \Pr[\mathcal{M}(D')\in\mathcal{O}]+\delta.
\end{align}
\end{definition}


Within our framework, we should consider the following guarantee of DP during forward propagation:
\begin{definition}[$(\epsilon, \delta)$-DP]\label{def:dp}
A randomized mechanism $\mathcal{M}$ is $(\epsilon,\delta)$-differentially private if for any two neighboring datasets $D$ and $D'$ that differ by one record, and we random sample a sample from both datasets, i.e., $x \in D, x' \in D'$, and any output $r \in \mathcal{O}$ of the output domain of $A$, we have the following.
\begin{align}\label{eq:dp-def}
    \Pr[\mathcal{M}(x)=r] \leq \exp(\epsilon) \Pr[\mathcal{M}(x')=r]+\delta.
\end{align}
\end{definition}

Such a guarantee is typically achieved by injecting noise into a function of the client’s data at a specific step, often at the cost of reduced utility—as demonstrated in our preliminary study in Section \ref{limitations}. To make the DP guarantee more practical, we explore whether probabilistic mask training can enable stronger privacy protection without increasing the noise level. Encouragingly, Bernoulli sampling aligns with this goal \cite{balle2018privacy, yang2022differentially, imola2021privacy, isik2022sparse}. In the following, we demonstrate how probabilistic mask training contributes to achieving DP guarantees in two key stages of our algorithm: the split learning step and the mask aggregation step. Following prior work \cite{thapa2022splitfed}, we analyze these stages separately due to their distinct data flows and independent noise injection mechanisms.



\begin{theorem}[Privacy Guarantee for Smashed Data Forwarding] \label{theorem:forward}
Before forwarding the smashed data to the server, a client adds Laplace noise $Lap(\Delta f/\epsilon)$ to the smashed data, where $\Delta f$ is the sensitivity of the function $f$ before applying a mask. By assuming that each parameter is sampled with a bounded probability $[c, 1-c]$, where $c \in (0, 0.5)$, this mechanism satisfies $(\epsilon_{amp}, 0)$-DP, where $\epsilon_{amp} = \ln \left((1 - c^d)\exp({\epsilon}) + c^d \right)$.
\end{theorem}

Prior to applying the masking strategy, the smashed data enjoys an $\epsilon$-DP guarantee due to the Laplace mechanism \cite{abadi2016deep}. The above theorem shows that integrating probabilistic masking results in a tighter privacy bound, with $\epsilon_{amp} < \epsilon$, thereby amplifying the differential privacy protection. 
The proof is in Appendix \ref{proof_theorem2}.

\begin{theorem}[Privacy Guarantee for Mask Aggregation] \label{theorem:backward}
Before mask aggregation as defined in Eq. \eqref{eq:mask_agg}, we consider two noise adding strategies as follows and provide DP guarantee without applying mask strategy for Bernoulli sampling:
\begin{itemize}[leftmargin=1em]
    \item \textbf{Adding noise to each local update:} Suppose the noise $z_{t, r}^{(k)} \sim \mathcal{N}(0, \sigma^2 \mathbf{I}_{d_b})$, where $\mathcal{N}(0, \sigma^2 \mathbf{I}_{d_b})$ is denoted by normal distribution with the mean of 0 and the variance of $\sigma^2 \mathbf{I}_{d_b}$. Suppose the update of the score mask is bounded by $\Gamma$. Therefore, 
    \begin{align}\label{eq:dp_local_update}
        \mathbf{s}_{t, r+1}^{(k)} &= \mathbf{s}_{t, r}^{(k)} - \frac{1}{|\mathcal{Q}_k|} \sum_{q \in \mathcal{Q}_k, \nabla q \in \nabla \mathcal{Q}_k} g_{t, r, q}^{(k)} + z_{t, r}^{(k)},
    \end{align}
    where
    \begin{align}
        g_{t, r, q}^{(k)} &= (\nabla q)^{\top} \left(\frac{\partial q}{\partial s}\right) / \max\left(1, \left\|(\nabla q)^{\top} \left(\frac{\partial q}{\partial s}\right)\right\|/\Gamma\right).
    \end{align}
    By setting $\sigma^2 \geq \frac{2 R \Gamma^2 \ln{(1/\delta)}}{\epsilon^2 |\mathcal{Q}_k|^2}$, this mechanism is $(\epsilon, \delta)$-DP. 
    \item \textbf{Adding noise to the mask sampling:} Suppose the noise $z_{t}^{(k)} \sim \mathcal{N}(0, \sigma^2 \mathbf{I}_{d_b})$, we further update the probabilistic mask for $\hat{\theta}_{t, R}^{(k)} = \text{clip}\left(\theta_{t, R}^{(k)}+z_{t}^{(k)}, c, 1-c\right)$. Therefore, a binary mask $\mathbf{M}_k$ is generated using $\mathbf{M}_k \sim \text{Bern}\left(\hat{\theta}_{t, R}^{(k)}\right)$. In this case, by setting $\sigma^2 \geq \frac{2 (1-2c)^2 \ln(1.25/\delta)}{\epsilon^2}$, we ensure the mechanism satisfies $(\epsilon, \delta)$-DP.
\end{itemize}
\end{theorem}

In the theorem above, we analyze how to inject noise with privacy budget $\epsilon$ prior to applying Bernoulli sampling.
As discussed in \citet{imola2021privacy}, if a mechanism satisfies $(\epsilon, \delta)$-differential privacy, applying Bernoulli sampling can further amplify its privacy guarantee. Thus, we can get amplified DP privacy budget $\epsilon_{amp}$ after Bernoulli sampling: $\epsilon_{amp} \leq \min_{\alpha > 1} \{\epsilon, \frac{d_b}{\alpha - 1} \cdot \log ((1-c)^{\alpha} c^{1-\alpha} + (1-c)^{\alpha} c^{1-\alpha})\}$, where $c\in (0,0.5)$ restricts the Bernoulli parameters to lie within the interval $[c,1-c]$. 
The detailed proof is provided in Appendix \ref{proof_theorem3}.

\section{Experiment} \label{sec:experiment}

\begin{table*}[t!]
    \centering
    \caption{Test accuracy under different experimental settings.}
\vspace{2px}
    \resizebox{\linewidth}{!}{
    \begin{tabular}{ccc||c|c|c|c|c|c|c}
         \toprule
         Data  & System  & Dataset & \multicolumn{3}{c|}{Training Parameters} & \multicolumn{4}{c}{Training Probabilistic Masks}  \\
         Hetero. & Hetero.  &   & Standalone & SplitFed & SplitFed-DP & SplitFed-PM & LG-FedAvg & DepthFL & \algo (ours)  \\
         \midrule
         Dirichlet & No & CIFAR-100 & 30.16 & \textbf{51.39} & 38.87 & 50.81 & \text{—} & \text{—} & 51.02  \\
         \midrule
         Personalized & No & FEMNIST & 74.31 &78.53 & 75.45 & 68.41 & 71.32 & \text{—} & \textbf{81.89}  \\
         & No & mmHAR & 80.69 & 85.48 & 82.86 & 77.21 & 83.76 & \text{—} & \textbf{91.18} \\
         & No & sonicHAR & 74.07 & 80.09 & 75.48 & 71.67 & 79.34  & \text{—} & \textbf{88.96} \\
         \midrule
         Dirichlet & Yes & CIFAR-100 & 16.72 & 27.09 & 21.15 & 14.06 & \text{—} & 31.21 & \textbf{42.47} \\
         Personalized & Yes & FEMNIST & 65.37 & 68.77 & 65.95 & 53.76 & \text{—} & 71.26 & \textbf{78.56} \\
         \bottomrule
         
    \end{tabular}
    }
    \label{tab:accu_result}
    \vspace{5px}
\end{table*}

We evaluate \algo across several datasets, data partitions, and backbones to answer the following questions:
\textbf{Q$1$:} How does probabilistic mask training affect accuracy and privacy in SFL?
\textbf{Q$2$:} Does the data-aware personalized mask design effectively cater to clients with unique data distributions?
\textbf{Q$3$:} How does the \algo perform across clients with varying computational resources?

\subsection{Datasets and Models} \label{sec:dataset_model}

\begin{table}[t]
    \centering
    \caption{Statistical information of datasets.}
\vspace{2px}
    \resizebox{\columnwidth}{!}{
    \begin{tabular}{lcccc}
    \hline
        Dataset & \# Clients & \makecell[c]{\# Samples\\ per client} & \# Classes & \makecell[c]{\# Data\\ partition} \\\hline
        CIFAR-100 & 100 & 600 & 100 & dirichlet  \\
        FEMNIST & 190 & 226 & 62 & by subject \\
        mmHAR & 10 & 204 & 17 & by subject \\
        sonicHAR & 10 & 204 & 17 & by subject \\\hline
    \end{tabular} 
    }
    \label{tab:datasets}
\vspace{5px}
\end{table}

 \noindent\textbf{Image Classification Datasets.} We begin with CIFAR-100 and FEMNIST for benchmark evaluation. CIFAR-100 \cite{krizhevsky2009learning} is split among $100$ clients using a Dirichlet distribution ($\alpha=0.3$) \cite{hsu2019measuring, jhunjhunwala2023fedexp}, with roughly $5{:}1$ train-test split per client. FEMNIST \cite{cohen2017emnist, caldas2018leaf} naturally exhibits non-IID data by assigning writers as clients. We sample $190$ writers with about $90\%$ of each client’s data used for training, reflecting realistic personalization settings.

\noindent\textbf{Real-world Evaluation (mmHAR \& sonicHAR)\footnote{All the data collection was approved by IRB of the authors’ institution.}.}
To validate the generalizability of our method across applications and modalities, we collected wireless sensing HAR data using a mmWave radar (TI AWR1843) and a smartphone (Samsung S9) with ultrasound sensing in realistic settings. Ten subjects performed 17 daily activities in three rooms, and data were manually labeled. Both devices sampled at 44.1 kHz. Radar signals were processed with FFTs to generate time–Doppler heatmaps \cite{rao2017introduction}, and ultrasound signals were demodulated with a coherent detector \cite{tse2005fundamentals} and transformed via STFT to obtain DFS profiles. Data were segmented into 3-second intervals, yielding 12 segments per class (9 for training, 3 for testing).

\noindent\textbf{Models.}
For the two image recognition datasets, we utilize a modified ResNet-$18$ architecture \cite{he2016deep}, where we replace the batch normalization (BN) layers with static BN counterparts \cite{diao2020heterofl} to conduct training. For the two HAR datasets, we employ a customized Transformer-based model \cite{wolf2019huggingface} comprising four identical layers. Each layer features $2$ attention heads with a model dimension of $768$ and a feed-forward network hidden dimension of $768 \times 2$.

\subsection{Experimental Setup}

\textbf{Baselines.}\quad To demonstrate the effectiveness of \algo, we compare \algo against the following baselines:

\begin{itemize}[leftmargin=1em]
    \item \textbf{Standalone:} In this method, each client independently trains its local bottom submodel using its own data, assisted by the server, without an aggregation step.
    \item \textbf{SplitFed \cite{thapa2022splitfed}:} This method represents the basic form of split federated learning as outlined in Section \ref{sec:SFL}.
    \item \textbf{SplitFed-DP \cite{thapa2022splitfed}:} This variant of SplitFed enhances privacy by injecting noise into communications from clients to the server, with a privacy budget of $0.1$. It corresponds to the minimum noise required to fully defend against data reconstruction attacks, based on carefully tuned evaluations across different privacy budgets.
    \item \textbf{SplitFed-PM:} This SplitFed variant replaces parameter training with probabilistic mask training, omitting data-aware personalization and adaptive splitting. It transmits probabilistic masks rather than binary masks for aggregation.
    \item \textbf{LG-FedAvg \cite{liang2020think}:} This approach enables layer-wise partial model personalization of the bottom submodel.
    \item \textbf{DepthFL \cite{kim2023depthfl}:} This technique employs self-distillation of knowledge among layers within bottom submodels to improve the training effectiveness of deeper layers.
\end{itemize}

\noindent\textbf{Data Heterogeneity Setup.}
To address Q$2$ posed at the beginning of Section \ref{sec:experiment}, we implement two data partitioning methods to study data heterogeneity. As detailed in Section \ref{sec:dataset_model}, we utilize a Dirichlet distribution with $\alpha = 0.3$ for CIFAR-100 and naturally partition FEMNIST and two HAR datasets by subject.

\noindent\textbf{System Heterogeneity Setup.}
For the ResNet-18 and Transformer models, each comprising four basic blocks, we set the splitting point just after the second basic block for all clients by default. To explore Q$3$, we vary the splitting layer position among clients, maintaining an equal ratio of $1:1:1:1$ for clients with the split positioned after the first, second, third, and fourth basic block, respectively.

\subsection{System Implementation}
In split federated learning, client contributions to the training process can be inconsistent due to technical constraints or user behavior. To simulate this variability, we randomly select $10\%$ of clients to participate in each communication round. However, for the two self-collected datasets, we include all clients in the training process. This decision is due to the limited number of clients available in these datasets.
We adopt FedAvg \cite{mcmahan2017communication} as our aggregation scheme, as it is the most widely used method in federated learning.

During the split federated learning process, each client performs $5$ local training epochs per communication round. The training runs for a total of $1000$, $500$, and $100$ rounds on the CIFAR-100, FEMNIST, and self-collected HAR datasets, respectively. The neural network is implemented using PyTorch and trained with the Adam optimizer. The learning rate is set at $10^{-3}$. Training occurs in batches of $32$ on a server equipped with an NVIDIA A6000 GPU and an Intel Xeon Gold $6245$ CPU. Inference tasks are carried out on the same machine. Each quantitative experiment is repeated three times using different random seeds, and the averaged results are reported in the subsequent results sections.

\begin{figure}[t]
    \centering
    \subfloat[Training Convergence.]{
        \includegraphics[width=0.48\linewidth]{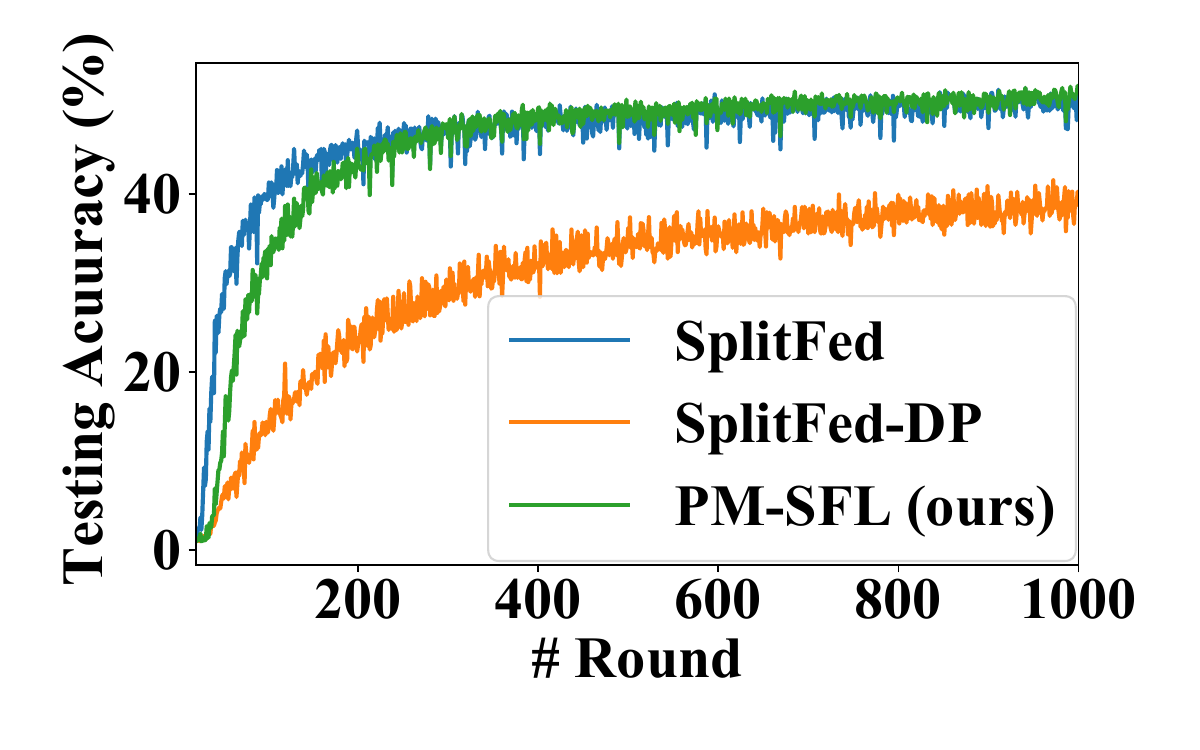} 
        \label{fig:exp_convergence_cifar}
    }
    \hfill
    \subfloat[Conmmunication Overhead.]{
        \includegraphics[width=0.48\linewidth]{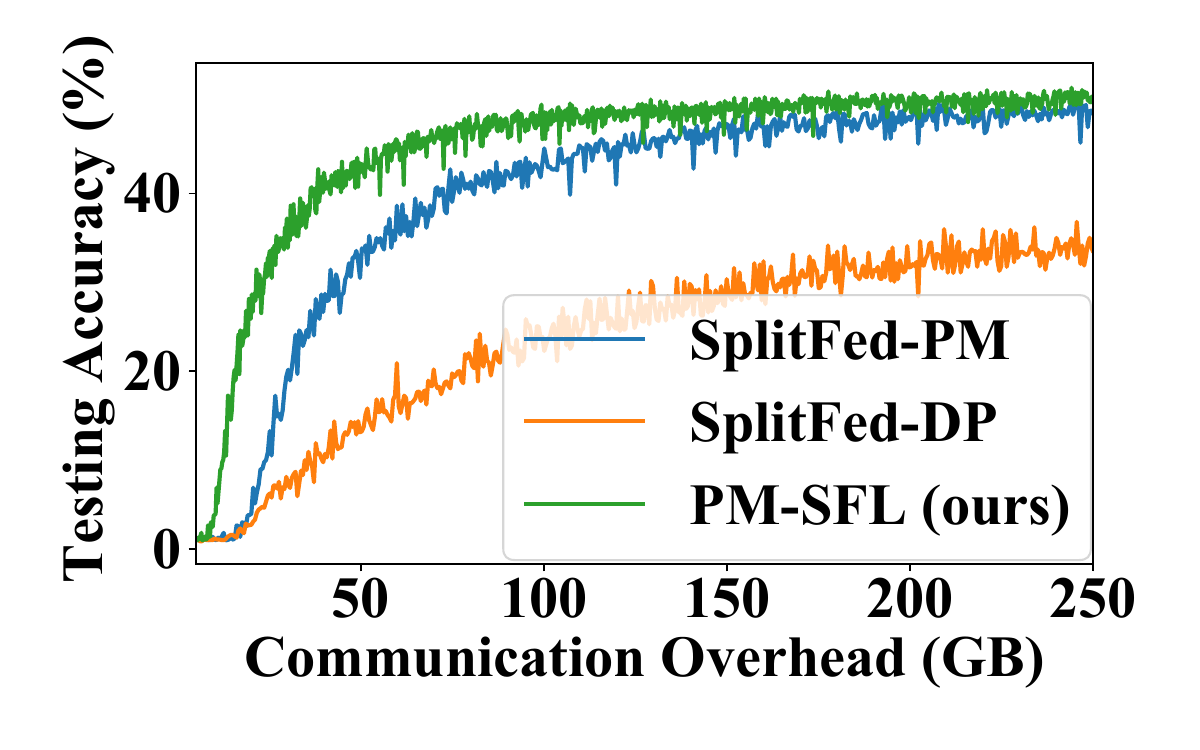} 
        \label{fig:exp_communication_cifar}
    }
    
    \caption{Performance of Probabilistic Mask Training on CIFAR-100 using ResNet-18.}
    \label{fig:exp_cifar}
\end{figure}

\subsection{Performance of Probabilistic Mask Training}

To address Q$1$, we compare \algo with baselines along two key dimensions: testing accuracy and data privacy. Testing accuracy is evaluated by averaging all clients' performance on their respective local datasets. 
For privacy evaluation, we follow the same reconstruction attack methodology described in Section \ref{limitations} and quantify reconstruction quality using the Structural Similarity Index Measure (SSIM), where lower values indicate stronger privacy protection.

\begin{figure}[t]
    \centering
    \includegraphics[width=0.95\linewidth]{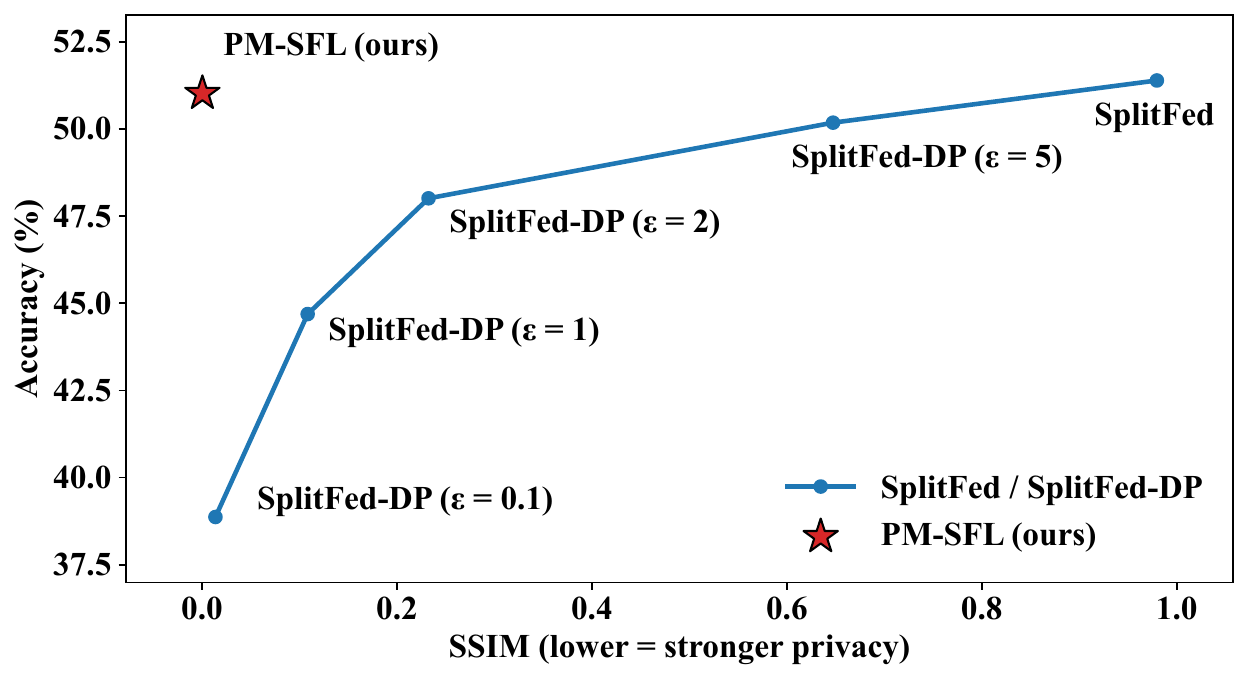} 
    
    \caption{Privacy--Accuracy Tradeoff Under Reconstruction Attacks on CIFAR-100 Using ResNet-18.}
    \label{fig:exp_SSIM_acc}
\end{figure}

We first provide a quantitative comparison of privacy and utility.
Figure \ref{fig:exp_SSIM_acc} summarizes the relationship between reconstruction similarity (SSIM) and classification accuracy for SplitFed and SplitFed-DP under different privacy budgets on CIFAR-100 using ResNet-18.
As the privacy budget decreases, SplitFed-DP achieves progressively lower SSIM, but this improvement comes at the cost of reduced classification accuracy, forming a clear privacy-accuracy tradeoff curve.
In contrast, \algo attains low SSIM without significantly sacrificing accuracy compared to the non-DP baseline, demonstrating that probabilistic mask training can provide strong privacy protection while maintaining model utility.

\begin{figure}[t]
    \centering
    \includegraphics[width=0.75\linewidth]{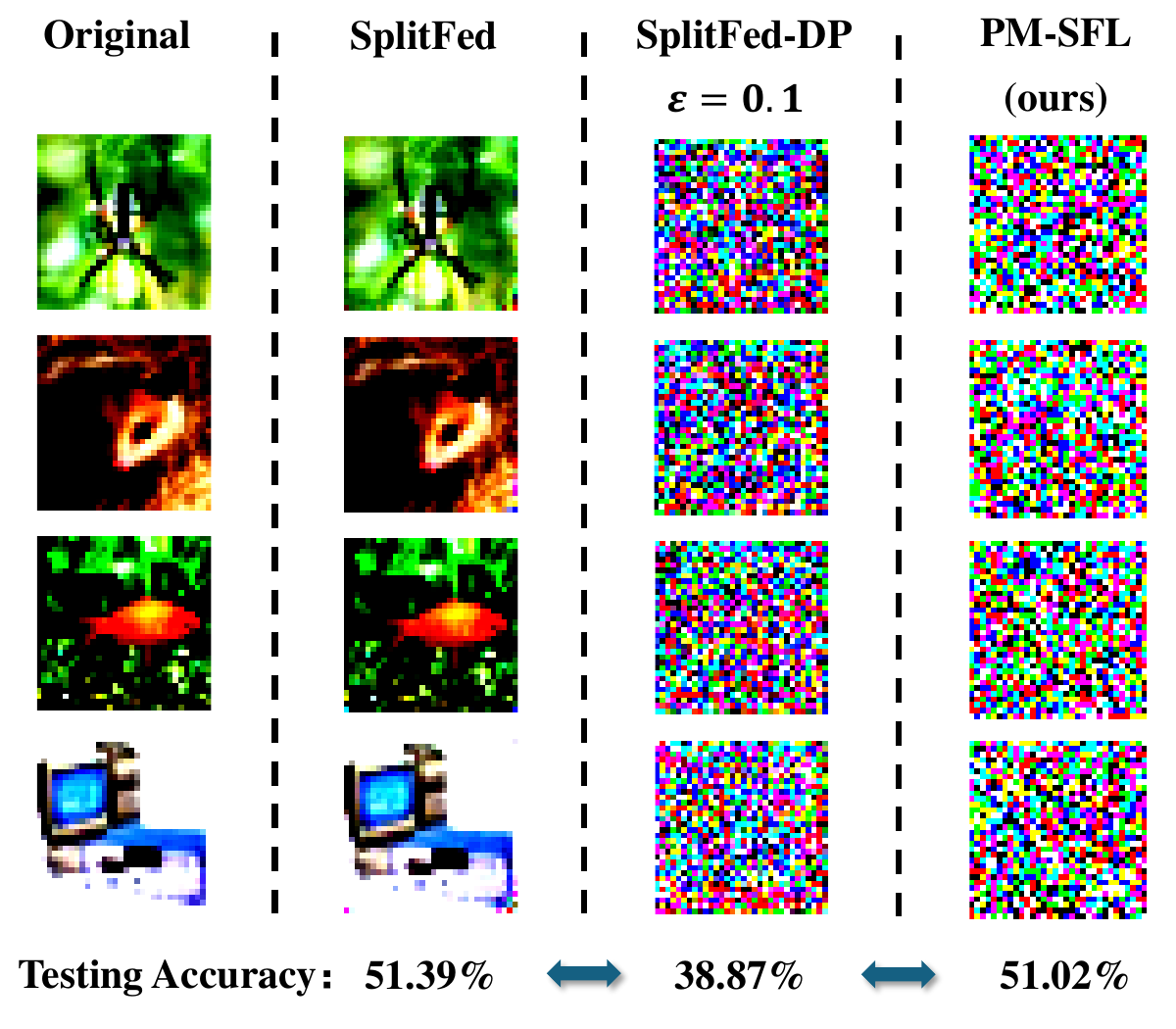} 

    \caption{Data Reconstruction Attack Results. }
    \label{fig:exp_privacy}
\end{figure}

To complement this quantitative analysis, we further present qualitative reconstruction results.
As shown in Figure \ref{fig:exp_privacy}, \algo significantly degrades the visual quality of reconstructed samples compared to SplitFed, consistent with its low SSIM values in Figure \ref{fig:exp_SSIM_acc}.
While SplitFed-DP with a strong privacy budget ($\epsilon = 0.1$) can also obscure reconstructions, it suffers from a substantial accuracy drop of approximately 12\%, highlighting the advantage of \algo in jointly preserving accuracy and privacy.

In addition, Figure \ref{fig:exp_convergence_cifar} and Figure \ref{fig:exp_communication_cifar} show that \algo maintains convergence behavior comparable to SplitFed while reducing communication overhead by transmitting binary masks instead of probabilistic masks, further improving training efficiency.

\subsection{Performance of Personalized Mask Training}

To address Q$2$, we conduct experiments on datasets that naturally support personalized data partitioning, where each client possesses unique written characters or performed activities. This setting poses a significant challenge for aggregation, as it requires effectively integrating highly personalized updates.

Probabilistic mask-based training, when applied without a personalization mechanism, suffers from severe performance degradation. As shown in the "SplitFed-PM" column in Table \ref{tab:accu_result}, its performance is even worse than the "Standalone" approach, where collaboration for bottom submodel training is entirely disabled. However, when incorporating our personalized mask design, the model achieves substantial performance improvements. Our fine-grained personalized mask selection method demonstrates a clear advantage over layer-level personalized mask selection, as represented by "LG-FedAvg", highlighting the effectiveness of the proposed approach in capturing client-specific adaptations.

\subsubsection{Hyper-parameter Tuning.}
There are two key hyper-parameters that require tuning in the proposed mask personalization method: the number of rounds needed to reach an initial agreement and the percentage of the mask to be personalized. We investigate these parameters to gain insights for optimization.

\begin{figure}[t]
    \centering
    \subfloat[Effects of Initial Agreement Rounds.]{
        \includegraphics[width=0.48\linewidth]{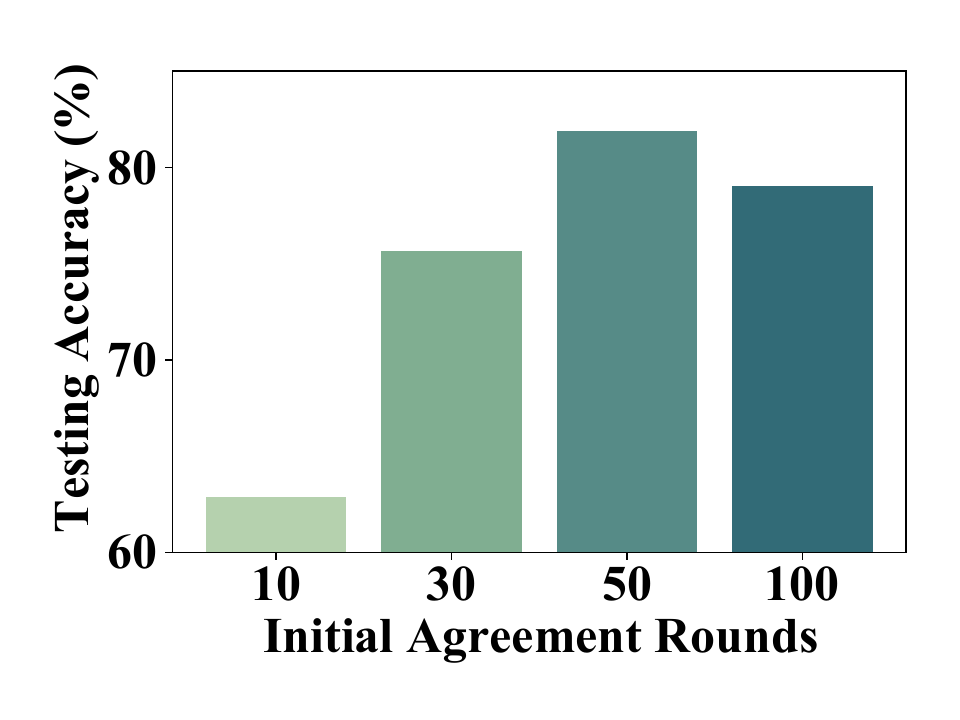} 
        \label{fig:exp_hyper_rounds}
    }
    \hfill
    \subfloat[Effects of Personalized Mask Ratio.]{
        \includegraphics[width=0.48\linewidth]{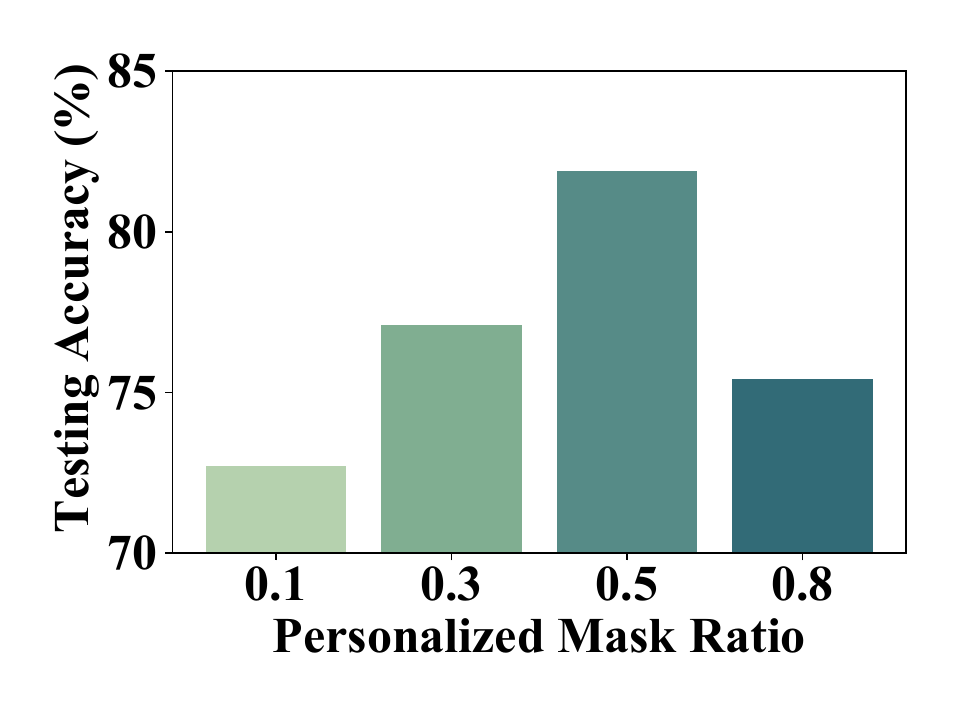} 
        \label{fig:exp_hyper_ratio}
    }
\vspace{2px}    
    \caption{Impact of Personalization Hyperparameters on \algo's Performance on FEMNIST with ResNet-18.}
    \label{fig:exp_hyper}
\end{figure}

\noindent\textbf{Initial Agreement Rounds.}\quad As described in Section \ref{sec:per_mask}, our personalized mask selection begins after clients reach an initial global consensus. Thus, determining the appropriate number of rounds for this initial agreement is crucial. Our experiments show that too few rounds lead to unstable training and poor convergence, while too many rounds slow down convergence and increase computational overhead, both resulting in suboptimal performance. As shown in Figure \ref{fig:exp_hyper_rounds}, $50$ rounds, equivalent to $10\%$ of total training rounds, yield the best performance.

\noindent\textbf{Personalized Mask Ratio.}\quad Another key hyperparameter to tune is the ratio of personalized masks to the total number of masks. This ratio plays a crucial role in balancing personalization and global knowledge sharing. As shown in Figure \ref{fig:exp_hyper_ratio}, our experimental results indicate that selecting half of the masks for personalization achieves the best overall performance.

\subsection{Performance of Knowledge Compensation}

To address Q$3$, we simulate scenarios where clients have varying local computational resources by adjusting the splitting layer position. As shown in Table \ref{tab:accu_result}, this model heterogeneity poses a significant challenge, regardless of the chosen backbone model, data distribution, or training paradigm (whether training parameters or probabilistic masks).

Existing approaches, such as "DepthFL" alleviate performance degradation caused by model heterogeneity. However, as discussed in Section \ref{sec:knoeledge_compensation}, these methods do not fully leverage the advantages of SFL. Our proposed approach enhances aggregation by integrating the knowledge learned on the top submodel, ensuring that all layers receive adequate training. As evidenced in Table \ref{tab:accu_result}, this knowledge compensation mechanism substantially improves testing accuracy under heterogeneous system settings, highlighting its effectiveness in facilitating adaptive model splitting.

\section{Related Work} \label{related_work}

\noindent\textbf{Mask-based Training.}\quad 
To the best of our knowledge, this is the first work to repurpose probabilistic masks for privacy-preserving Split Federated Learning (SFL). Prior binary masking methods, such as FedMask \cite{li2021fedmask} in standard FL, generate fixed binary masks through deterministic thresholding, which lack structured randomness and are therefore more vulnerable to privacy leakage and reconstruction attacks. In contrast, our probabilistic masking leverages Bernoulli sampling to introduce structured randomness, providing stronger privacy guarantees and smoother optimization.

\noindent\textbf{Protecting Against Privacy Leakage.}\quad Among defenses for data reconstruction attacks in FL, encryption-based methods, such as secure multi-party computation (SMC) \cite{bonawitz2017practical,lia2020privacy} and homomorphic encryption (HE) \cite{kim2018efficient,yue2023gradient}, provide strong theoretical guarantees but introduce high computational and communication overhead, increased storage needs, and require changes to the FL architecture. Perturbation-based defenses \cite{gao2021privacy,gao2023automatic,huang2020instahide,wu2024concealing,zhang2017mixup} inject noise into inputs, gradients, or training processes, often at the cost of degraded model performance. Pruning-based approaches \cite{zhu2019deep, sun2021soteria, xue2024revisiting} reduce leakage by masking gradients based on magnitude but typically lead to information loss and are infeasible for resource-constrained devices due to the need to load the full model. 
In SL, Patch Shuffling \cite{yao2022privacy} is effective for vision tasks with spatial priors but not applicable to other modalities like time-series data (e.g., mmHAR \& sonicHAR in our experiments). SplitGuard \cite{erdogan2022splitguard} detects malicious clients through gradient analysis but addresses a different threat model and does not prevent smashed data leakage. NoPeek \cite{vepakomma2020nopeek} reduces input–activation correlation via regularization but incurs overly high training time, scales poorly with batch size.

\noindent\textbf{Data Heterogeneity.}\quad 
Personalized federated learning addresses data heterogeneity by tailoring local models to their specific data distributions \cite{tan2022towards, wang2024towards}. We focus on partial model personalization, which updates only parts of the model to better fit local tasks \cite{collins2021exploiting, li2020lotteryfl, mugunthan2022fedltn, tamirisa2024fedselect, liang2020think}. This avoids overfitting and preserves global knowledge \cite{kirkpatrick2017overcoming, pillutla2022federated}. However, existing methods lack tailored optimization for probabilistic masking, limiting both training efficiency and model accuracy. 
FedMask \cite{li2021fedmask} applies partial model personalization to binary mask training, but it fixes which mask elements are updated through one-shot pruning, permanently discarding others and limiting adaptability.

\noindent \textbf{System Heterogeneity.}\quad
Clients in SFL often have diverse computational resources. A common solution tailors model sizes to client capacity \cite{lin2020ensemble, itahara2021distillation, zhang2023distill, zhang2021parameterized}, often assuming access to public data, which is unavailable in domains like wireless sensing. Adaptive model-splitting approaches, more suitable for SFL, reduce client load but face performance degradation because clients with only shallow layers contribute little to the training of deeper layers, which is a challenge that tiering-based strategies cannot address \cite{mohammadabadi2024speed}. Although other FL methods \cite{kim2023depthfl, diao2020heterofl, ilhan2023scalefl, wu2024fiarse} can be incorporated, they miss the opportunity for server-side knowledge compensation that is crucial in adaptive model-splitting SFL.

\noindent\textbf{Additional Adversarial Scenarios.}\quad 
While the primary focus of \algo is privacy preservation against reconstruction attacks, its architectural design also offers inherent robustness to other adversarial threats, including model poisoning \cite{bagdasaryan2020backdoor, fang2020local} and malicious mask manipulation \cite{yan2024skymask}. In particular, the stochasticity introduced by Bernoulli sampling functions as a decentralized defense mechanism by attenuating the deterministic influence of any single client on the global mask estimator. Although a comprehensive empirical evaluation under active poisoning attacks or adaptive adversaries is beyond the scope of this work, we identify these scenarios as important directions for future investigation.




\section{Conclusion} \label{conclusion}

This paper presents \algo, a novel Split Federated Learning framework designed to address the critical challenges of privacy preservation, data heterogeneity, and system heterogeneity in federated settings. By leveraging probabilistic mask training, \algo introduces structured randomness that protects client data from reconstruction attacks while avoiding the performance degradation common in noise-injection-based defenses. To support diverse client needs, \algo further incorporates data-aware personalized mask learning and layer-wise knowledge compensation, enabling efficient training across heterogeneous data distributions and diverse client capabilities. Our theoretical and empirical results demonstrate that \algo improves accuracy, communication efficiency, and robustness to privacy attacks, highlighting its potential to advance the scalability and security of SFL in real-world deployments.

\begin{acks}
This work was supported by the U.S. National Science Foundation under Grants IIS-2141037 and CNS-2154059.
\end{acks}

\bibliographystyle{ACM-Reference-Format}
\balance
\bibliography{reference}

\clearpage
\allowdisplaybreaks
\appendix

\section{Backward Propagation Derivation} \label{sec:backward_derivation}

\noindent \textbf{Step $2.3$ (Local Backward):} After receiving the gradient w.r.t. the smahsed data, client $k \in K_t$ updates its local score $\mathbf{s}_{t, r}^{(k)}$, which follows 
\begin{align}
    \mathbf{s}_{t, r+1}^{(k)} = \mathbf{s}_{t, r}^{(k)} - \frac{1}{|\mathcal{Q}_k|} \sum_{q \in \mathcal{Q}_k, \nabla q \in \nabla \mathcal{Q}_k} (\nabla q)^{T} \left(\frac{\partial q}{\partial s}\right),
\end{align}
where
\begin{align}
    \frac{\partial q}{\partial s} &= \frac{\partial f_{\mathbf{w}_b \odot \mathbf{M}_k}(x)}{\partial \mathbf{w}_b \odot \mathbf{M}_k} \frac{\partial \mathbf{w}_b \odot \mathbf{M}_k}{\partial s} \\
    &=\frac{\partial f_{\mathbf{w}_b \odot \mathbf{M}_k}(x)}{\partial \mathbf{w}_b \odot \mathbf{M}_k} \frac{\partial \mathbf{w}_b \odot \mathbf{M}_k}{\partial \sigma(s)} \frac{\partial \sigma(s)}{\partial s}\\
    &= \frac{\partial f_{\mathbf{w}_b \odot \mathbf{M}_k}(x)}{\partial \mathbf{w}_b \odot \mathbf{M}_k} \odot \mathbf{w}_b \odot \sigma(s) \odot \sigma'(s).
\end{align}
$\sigma'(\cdot)$ is the derivative of $\sigma(\cdot)$. Following \cite{isik2022sparse}, the last equality is based on the straight-through estimator \cite{bengio2013estimating}, which considers the Bernoulli sampling operation differentiable during backward propagation, simply equal to the probability mask $\sigma(s)$.

\section{Potential Leakage when Transmitting Probabilistic Mask} \label{sec:leakage_analysis}



Consider a local network modeled as a multilayer perceptron (MLP) with $m$ sequential dense layers. For each layer $i$ (where $i = 1, \dots, m$), the output is computed as
\begin{align}
y_i = (W_{i} \odot M_i) x_i,
\end{align}
with the next layer’s input defined as $x_{i+1} = y_i$. Here, $W_i \in \mathbb{R}^d$ represents the weight matrix, and the binary mask $M_i \in \{0, 1\}^d$ is sampled from $\sigma(s_i)$.

In this framework, the client sends the final output $y_m$ to the server, which continues the forward propagation and trains its model accordingly. As described in Step 2.2 of Section~\ref{sec:prob_mask}, the server returns the gradient of the loss $\ell$ with respect to $y_m$, denoted as $\frac{\partial \ell}{\partial y_m}$.

According to Eq.~\eqref{eq:local_update}, the gradients are computed as follows:

\begin{itemize}
    \item \textbf{Gradient with respect to the mask parameter $s_m$:}
    \begin{align}
    \frac{\partial \ell}{\partial s_m} = \frac{\partial \ell}{\partial y_m} \cdot x_m \cdot W_m \cdot \sigma(s) \cdot \sigma'(s)
    \end{align}
    
    \item \textbf{Gradient with respect to the input $x_m$:}
    \begin{align}
    \frac{\partial \ell}{\partial x_m} = \frac{\partial \ell}{\partial y_m} \cdot (W_m \odot M_{m})
    \end{align}
\end{itemize}

Assume that for every layer $i \in \{1, \dots, m\}$, the gradient $\frac{\partial \ell}{\partial s_i}$ is visible to the server. With this information:

\begin{itemize}
    \item From the expression for $\frac{\partial \ell}{\partial s_m}$, one can recover the value of $x_m$ (which is equivalent to $y_{m-1}$).
    \item Knowing that $y_m = (W_m \odot M_m) x_m$ is also visible allows the computation of $\frac{\partial \ell}{\partial x_m}$, which is the same as $\frac{\partial \ell}{\partial y_{m-1}}$.
\end{itemize}

By iteratively applying this reasoning across the layers, the server could ultimately recover the raw input $x_1$.

In summary, when the updates for the score mask $\mathbf{s}$ (i.e., the gradients $\frac{\partial \ell}{\partial s_i}$ for all layers) are visible to the server, the risk of data reconstruction is higher compared to transmitting only a sampled mask. 

\section{Proof of Theorem \ref{theorem:data_reconstruction}} \label{proof_theorem1}

A trivial lower bound is simply 0—if there’s any chance that the guess is perfect (i.e. \(\hat{M}=M\) with nonzero probability), then on those events the error is zero and hence
\begin{align}
\mathbb{E}\|x-\hat{x}\| \ge 0.
\end{align}

However, if we assume that every \(\hat{M}\) in the support of \(P(\hat{M})\) leads to a nonzero deviation (or if we restrict attention to cases where \(\hat{M}\neq M\)), we can obtain a nontrivial lower bound by working in terms of the singular values of the involved matrices. Recall that
\begin{align}
x-\hat{x} = (W\odot \hat{M})^{-1}\bigl[W\odot (\hat{M}-M)x\bigr].
\end{align}
Using the property that for any matrix \(A\) we have
\begin{align}
\|Ax\| \ge \sigma_{\min}(A)\|x\|,
\end{align}
we may write
\begin{align}
\|x-\hat{x}\| \ge \sigma_{\min}\Bigl((W\odot \hat{M})^{-1}(W\odot (\hat{M}-M))\Bigr)\|x\|.
\end{align}

Because the minimum singular value of a product of matrices is bounded by the product of their minimum singular values (and noting that for an invertible matrix \(A\), \(\sigma_{\min}(A^{-1}) = 1/\sigma_{\max}(A)\)), we get
\begin{align}
&\quad \sigma_{\min}\Bigl((W\odot \hat{M})^{-1}(W\odot (\hat{M}-M))\Bigr)\\
&\ge \sigma_{\min}((W\odot \hat{M})^{-1})\,\sigma_{\min}(W\odot (\hat{M}-M))\\
&=\frac{\sigma_{\min}(W\odot (\hat{M}-M))}{\sigma_{\max}(W\odot \hat{M})}.
\end{align}
Thus, for each \(\hat{M}\) we have
\begin{align}
\|x-\hat{x}\| \ge \frac{\sigma_{\min}(W\odot (\hat{M}-M))}{\sigma_{\max}(W\odot \hat{M})}\|x\|.
\end{align}

Taking the expectation (over those \(\hat{M}\) that always yield a nonzero error) yields
\begin{align}
\mathbb{E}\|x-\hat{x}\| \ge \sum_{\hat{M}} P(\hat{M})\,\frac{\sigma_{\min}(W\odot (\hat{M}-M))}{\sigma_{\max}(W\odot \hat{M})}\|x\|.
\end{align}

\section{Proof of Theorem \ref{theorem:forward}} \label{proof_theorem2}

Suppose we have a single dense layer with parameter \(\mathbf{w}_b \in \mathbb{R}^{d}\), so that the function is defined by 
\begin{align}
f_{\mathbf{w}_b}: \mathbb{R}^d \to \mathbb{R}.
\end{align}
Before forwarding the smashed data to the server, a client applies a binary mask \(M\in\{0,1\}^{d}\) (with each coordinate sampled with a probability in \([c,1-c]\), where \(c\in (0,0.5)\)) and then adds Laplace noise \(z\sim Lap(\Delta f/\epsilon)\), resulting in 
\begin{align}
f_{M}(x) = (\mathbf{w}_b \odot M)\cdot x + z,
\end{align}
with the sensitivity defined as 
\begin{align}
\Delta f = \max_{x,x'}\|\mathbf{w}_b (x-x')\|.
\end{align}
Implicitly, we assume there is an input data $x = \boldsymbol{0}$. Our goal is to show that this mechanism satisfies \((\epsilon_{amp},0)\)-DP with 
\begin{align}
\epsilon_{amp} = \ln\Bigl((1-c^d)\exp(\epsilon) + c^d\Bigr).
\end{align}

We begin with the following lemmas.

\begin{lemma}[\cite{dwork2014algorithmic}]
A mechanism that adds Laplace noise \(z\sim Lap(\Delta f/\epsilon)\) to a function \(f\) satisfies \((\epsilon,0)\)-DP.
\end{lemma}

\begin{lemma}
For any fixed mask \(M\in\{0,1\}^{d}\), the mechanism defined by
\begin{align}
f_{M}(x) = (\mathbf{w}_b \odot M)\cdot x + z,\quad z\sim Lap(\Delta f/\epsilon),
\end{align}
satisfies \((\epsilon,0)\)-DP.
\end{lemma}

\begin{lemma}\label{lemma:3}
For any mask \(M\in\{0,1\}^{d}\) and any output value \(u\), the following holds:
\begin{align}
P\Bigl(f_{\boldsymbol{0}}(x)=u\Bigr) \geq \exp(-\epsilon)\, P\Bigl(f_{M}(x)=u\Bigr),
\end{align}
where \(f_{\boldsymbol{0}}(x)\) denotes the mechanism with the zero mask (i.e. \(M=\boldsymbol{0}\)).
\end{lemma}

Based on these lemmas, we now consider the overall mechanism that averages the output over the distribution of masks. For any fixed output \(u\), denote
\begin{align}
P_{(M,x)}(u) \triangleq P\Bigl(f_M(x)=u\Bigr),
\end{align}
and let \(P(M)\) be the probability of a mask \(M\). Our goal is to show that there exists an \(\epsilon'\) (to be identified as \(\epsilon_{amp}\)) such that for any pair of neighboring inputs \(x\) and \(x'\),
\begin{align}
\sum_{M}P(M)\,P_{(M,x)}(u) \le \exp(\epsilon') \sum_{M}P(M)\, P_{(M,x')}(u).
\end{align}

We begin by splitting the summation into the case when the mask is \(\boldsymbol{0}\) and when it is nonzero:
\begin{align}
\sum_{M}P(M)\,P_{(M,x)}(u)
= P(\boldsymbol{0})\,P_{(\boldsymbol{0},x)}(u) + \sum_{M\neq \boldsymbol{0}}P(M)\,P_{(M,x)}(u).
\end{align}
Since, by the above lemmas, the mechanism with any fixed mask satisfies \((\epsilon,0)\)-DP, we have
\begin{align}
P_{(\boldsymbol{0},x)}(u) \le \exp(\epsilon) \, P_{(\boldsymbol{0},x')}(u),
\end{align}
and similarly (using Lemma~\ref{lemma:3}) 
\begin{align}
P_{(M,x)}(u) \le \exp(\epsilon)\, P_{(M,x')}(u),\quad\forall\, M\neq \boldsymbol{0}.
\end{align}
Thus, we obtain
\begin{align}
&\sum_{M}P(M)\,P_{(M,x)}(u) \\
\le &P(\boldsymbol{0})\,P_{(\boldsymbol{0},x')}(u) + \exp(\epsilon)\sum_{M\neq \boldsymbol{0}}P(M)\,P_{(M,x')}(u).
\end{align}
Define
\begin{align}
a = P(\boldsymbol{0})\,P_{(\boldsymbol{0},x')}(u),\quad b = \sum_{M\neq \boldsymbol{0}}P(M)\,P_{(M,x')}(u).
\end{align}
Then the inequality becomes
\begin{align}
a + \exp(\epsilon)b \le \exp(\epsilon')(a+b),
\end{align}
with the smallest possible \(\epsilon'\) satisfying the equality
\begin{align}
a + \exp(\epsilon)b = \exp(\epsilon')(a+b).
\end{align}
This equality yields
\begin{align}
\exp(\epsilon') = 1 + \frac{\exp(\epsilon)-1}{1+\frac{a}{b}}.
\end{align}

Next, we bound the ratio \(a/b\). Note that by Lemma~\ref{lemma:3}, for any nonzero mask \(M\),
\begin{align}
P_{(\boldsymbol{0},x')}(u) \ge \exp(-\epsilon)\,P_{(M,x')}(u).
\end{align}
Hence,
\begin{align}
b &= \sum_{M\neq \boldsymbol{0}}P(M)\,P_{(M,x')}(u)\\
&\le \exp(\epsilon)P_{(\boldsymbol{0},x')}(u) \sum_{M\neq \boldsymbol{0}}P(M)\\
&=\exp(\epsilon) \,P_{(\boldsymbol{0},x')}(u)\left(1-P(\boldsymbol{0})\right).
\end{align}
It follows that
\begin{align}
\frac{a}{b} \ge \frac{P(\boldsymbol{0})\,P_{(\boldsymbol{0},x')}(u)}{\exp(\epsilon) \,P_{(\boldsymbol{0},x')}(u)(1-P(\boldsymbol{0}))}
=\frac{P(\boldsymbol{0})}{\exp(\epsilon)(1-P(\boldsymbol{0}))}.
\end{align}

Since each coordinate of the mask is sampled with probability at least \(c\), the probability that \(M\) is the zero mask satisfies
\begin{align}
P(\boldsymbol{0}) \ge c^d.
\end{align}
Thus,
\begin{align}
\frac{a}{b} \ge \frac{c^d}{\exp(\epsilon)(1-c^d)}.
\end{align}
Plugging this into the expression for \(\exp(\epsilon')\), we have
\begin{align}
\exp(\epsilon') = 1 + \frac{\exp(\epsilon)-1}{1 + \frac{a}{b}}
\le 1 + \frac{\exp(\epsilon)-1}{1 + \frac{c^d}{1-c^d}}
= (1-c^d)\exp(\epsilon) + c^d.
\end{align}
Taking logarithms on both sides yields
\begin{align}
\epsilon_{amp} \le \ln\Bigl((1-c^d)\exp(\epsilon) + c^d\Bigr).
\end{align}
This completes the proof that the mechanism satisfies \((\epsilon_{amp},0)\)-DP with 
\begin{align}
\epsilon_{amp} = \ln\Bigl((1-c^d)\exp(\epsilon) + c^d\Bigr).
\end{align}

\section{Proof of Theorem \ref{theorem:backward}} \label{proof_theorem3}

We consider two different mechanisms: one in which noise is added to each local update and one in which noise is added to the probabilistic mask before Bernoulli sampling. In both cases, we use the standard Gaussian mechanism and the post–processing property of differential privacy (DP).

\subsection{Adding Noise to Local Updates}

Each client’s update is given by:
\begin{align}
\mathbf{s}_{t, r+1}^{(k)} = \mathbf{s}_{t, r}^{(k)} - \frac{1}{|\mathcal{Q}_k|} \sum_{q\in \mathcal{Q}_k} g_{t, r, q}^{(k)} + z_{t, r}^{(k)},
\end{align}
where each gradient \(g_{t, r, q}^{(k)}\) is clipped such that \(\|g_{t, r, q}^{(k)}\| \le \Gamma\) and the noise \(z_{t, r}^{(k)}\) is drawn from
\begin{align}
z_{t, r}^{(k)} \sim \mathcal{N}(0,\, \sigma^2 \mathbf{I}_{d_b}).
\end{align}

\subsubsection{Single-Iteration Privacy:}
The sensitivity of the averaged gradient is bounded by
\begin{align}
\Delta \le \frac{\Gamma}{|\mathcal{Q}_k|}.
\end{align}
Thus, by the standard Gaussian mechanism, adding noise in a single iteration ensures \((\epsilon_0,\delta_0)\)-DP if
\begin{align}
\sigma \ge \frac{(\Gamma/|\mathcal{Q}_k|)\sqrt{2\ln(1.25/\delta_0)}}{\epsilon_0}.
\end{align}

\subsubsection{Composition Over \(R\) Iterations:}
Since the client performs \(R\) iterations, we must compose the privacy losses. Using advanced composition (or a moment accountant analysis), the overall privacy loss roughly scales as
\begin{align}
\epsilon \approx \sqrt{2R\ln\left(\frac{1}{\delta}\right)}\, \epsilon_0,
\end{align}
with \(\delta\) being the overall target failure probability (noting that \(\delta \ge R\,\delta_0\) in many cases).

Substituting the per-iteration guarantee into the composition yields a condition on the noise level such that
\begin{align}
\sigma \gtrsim \frac{\Gamma\sqrt{2R\ln(1/\delta)}}{|\mathcal{Q}_k|\epsilon}.
\end{align}
Squaring both sides, we obtain the sufficient condition for ensuring \((\epsilon,\delta)\)-DP:
\begin{align}
\sigma^2 \ge \frac{2 R \Gamma^2 \ln(\delta^{-1})}{\epsilon^2 |\mathcal{Q}_k|^2}.
\end{align}
This is the condition stated in the theorem for noise added to local updates.

\subsection{Adding Noise to Mask Sampling}

In the second strategy, the probabilistic mask is computed by first perturbing the output of the sigmoid function:
\begin{align}
\hat{\theta}_{t, R}^{(k)} = \operatorname{clip}\Big(\theta_{t, R}^{(k)} + z_{t}^{(k)},\, c,\, 1-c\Big),
\end{align}
where
\begin{align}
z_{t}^{(k)} \sim \mathcal{N}(0,\, \sigma^2 \mathbf{I}_{d_b}).
\end{align}
After this perturbation and clipping, the binary mask is generated as
\begin{align}
\mathbf{M}_k \sim \operatorname{Bernoulli}\left(\hat{\theta}_{t, R}^{(k)}\right).
\end{align}

Because the Bernoulli sampling is a post–processing of \(\hat{\theta}_{t, R}^{(k)}\), the overall privacy guarantee is determined by the Gaussian noise added to \(\theta_{t, R}^{(k)}\). Due to the clipping to the interval \([c, 1-c]\), the sensitivity of the function
\begin{align}
f(\mathbf{s}_{t, R}^{(k)}) = \operatorname{clip}\Big(\sigma(\mathbf{s}_{t, R}^{(k)}),\, c,\, 1-c\Big)
\end{align}
is bounded by
\begin{align}
\Delta \le 1-2c,
\end{align}
since the output range is at most \(1-c - c = 1-2c\).

Thus, by the Gaussian mechanism, the mechanism is \((\epsilon,\delta)\)-DP provided that
\begin{align}
\sigma \ge \frac{(1-2c)\sqrt{2\ln(1.25/\delta)}}{\epsilon},
\end{align}
or equivalently,
\begin{align}
\sigma^2 \ge \frac{2 (1-2c)^2 \ln(1.25/\delta)}{\epsilon^2}.
\end{align}
This is precisely the condition given in the theorem for noise added to mask sampling.

\subsection{Summary and Conclusion}

To summarize, we have shown that:
\begin{itemize}
    \item \textbf{Adding noise to each local update:} The mechanism satisfies \((\epsilon,\delta)\)-DP provided that
    \begin{align}
    \sigma^2 \ge \frac{2 R c^2 \ln(\delta^{-1})}{\epsilon^2 |\mathcal{Q}_k|^2}.
    \end{align}
    \item \textbf{Adding noise to mask sampling:} The mechanism satisfies \((\epsilon,\delta)\)-DP provided that
    \begin{align}
    \sigma^2 \ge \frac{2 (1-2c)^2 \ln(1.25/\delta)}{\epsilon^2}.
    \end{align}
\end{itemize}

These results follow from applying the standard Gaussian mechanism, combined with the composition theorem (for the local update case) and the post–processing property (for the mask sampling case). Hence, the derived privacy guarantees are consistent with the statement of Theorem~\ref{theorem:backward}.

\end{document}